\documentclass[10pt,twocolumn,letterpaper]{article}

\usepackage{cvpr} % uncomment this for the final submission
\usepackage{times}
\usepackage{epsfig}
\usepackage{graphicx}
\usepackage{amsmath}
\usepackage{amssymb}
\usepackage{xcolor}
\usepackage{siunitx}
\usepackage[pagebackref=true,breaklinks=true,colorlinks,bookmarks=false]{hyperref}

\begin{document}

%%%%%%%%% TITLE
\title{PIU: Proximity-guided Identity Unlearning in ID-Conditioned Diffusion Models}

% Authors: Edgar, Mauro, Žiga, Vito, Peter, Darian
% \and

\author{Jose Edgar Hernandez Cancino Estrada$^{1,}$\thanks{Jose Edgar Hernandez Cancino Estrada and Mauro Díaz Lupone are first authors with equal contributions.}\;, Mauro Díaz Lupone$^{1,*}$, Žiga Emeršič$^{1}$, \\ Vitomir Štruc$^{2}$, Peter Peer$^{1}$, and Darian Tomašević$^{1}$ \\
$^{1}$ University of Ljubljana, Faculty of Computer and Information Science, Ljubljana, Slovenia\\
$^{2}$ University of Ljubljana, Faculty of Electrical Engineering, Ljubljana, Slovenia \\
{\tt\small edgarcancinoe@exatec.tec.mx, maurodiazlupone@gmail.com, ziga.emersic@fri.uni-lj.si,} \\{\tt\small vitomir.struc@fe.uni-lj.si, \{peter.peer, darian.tomasevic\}@fri.uni-lj.si}
\vspace{-4mm}
}

\maketitle
\thispagestyle{empty}

%%%%%%%%% ABSTRACT

\begin{abstract}
% Identity-conditioned diffusion models enable high-quality and identity-consistent face generation, but in privacy-sensitive domains such as human faces they also raise serious concerns, as models may continue to synthesize a person even when that identity should be removed.
Identity-conditioned diffusion models enable high-quality and identity-consistent face generation, but they also raise severe privacy concerns, as models may continue to synthesize individuals despite their right to be forgotten.
While machine unlearning has been extensively studied for concept and data removal, identity unlearning remains largely unexplored, particularly in models conditioned directly on identity embeddings rather than text prompts. In this work, we study identity unlearning in Arc2Face, a state-of-the-art identity-conditioned latent diffusion model for face generation, and introduce \textbf{P}roximity-guided \textbf{I}dentity \textbf{U}nlearning (\textbf{PIU}), an anchor-guided framework for identity unlearning. Specifically, we formulate identity removal as an identity replacement objective that reassigns the source identity to a selected anchor identity in the learned identity space, and we complement it with a proximity-based anchor selection strategy motivated by the geometry of ArcFace representations. We further show that effective unlearning can be achieved through localized fine-tuning of a small subset of identity-sensitive cross-attention layers. Experiments across multiple target identities show that our framework effectively suppresses generation of the target identity while preserving realism and identity consistency for retained identities, as validated by improved performance on unlearning and image-quality metrics, together with qualitative evaluation. The source code for the PIU identity unlearning framework is made publicly available at \href{https://github.com/edgarcancinoe/piu-unlearning}{https://github.com/edgarcancinoe/piu-unlearning}.  %\url{https://github.com/edgarcancinoe/piu_unlearning}.
\end{abstract}

%%%%%%%%% BODY TEXT
\vspace{-4mm}
\section{Introduction}
\label{sec:introduction}

\begin{figure}
    \centering
    \small
    \begin{tabular}{@{}p{0.22\linewidth}@{}p{0.23\linewidth}@{}p{0.22\linewidth}@{}}
        \centering Target ID & \centering Anchor ID & \centering Unlearned ID
    \end{tabular}
    \includegraphics[width=0.66\linewidth, trim={1.8 0 0 11}, clip]{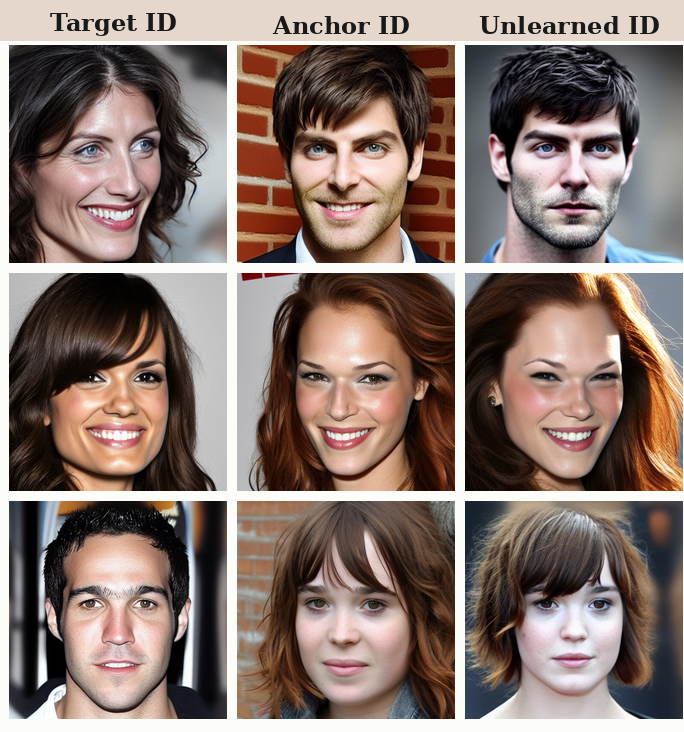}
    %\vspace{-4mm}
    \caption{\textbf{PIU unlearns a target identity from an identity-conditioned diffusion model~\cite{papantoniou2024arc2face} by mapping it to a selected anchor.} Afterwards, conditioning on the target identity produces images aligned with the anchor rather than the original subject.}
    \label{fig:manual_identity_grid}
    \vspace{-4mm}
\end{figure}

Diffusion models are nowadays the dominant paradigm for high-quality image generation, serving as the foundation for many state-of-the-art conditional generative systems~\cite{rombach2022high, ruiz2023dreambooth, papantoniou2024arc2face}. In biometrics, identity-driven synthesis has garnered particular attention for creating synthetic face images suitable for privacy-aware training and evaluation of deep models~\cite{boutros2023synthetic}. While text-based personalization methods can adapt pretrained generative backbones to specific subjects~\cite{ruiz2023dreambooth}, their textual conditioning often compromises identity consistency across variations in pose, illumination, and expression~\cite{tomavsevic2025id}. In contrast, identity-conditioned models like Arc2Face~\cite{papantoniou2024arc2face} directly condition on rich identity features of pretrained recognition models, enabling identity-consistent yet diverse synthesis. %~\cite{papantoniou2025id, di2026arc2morph}. 
Despite their value in producing large-scale synthetic datasets, such models also amplify privacy and misuse risks, including deepfake abuse~\cite{yan2023ucf}, which is reinforced by evidence that diffusion models can memorize and reproduce training data~\cite{carlini2023extracting,somepalli2023diffusion,somepalli2023understanding}. These vulnerabilities, coupled with regulations like the CCPA~\cite{goldman2020introduction} and GDPR~\cite{protection2018general} that formalize rights to erasure, necessitate the development of techniques to remove specific information from trained models without full retraining. This establishes a challenging objective: making ID-conditioned models forget a target individual.

To this end, machine unlearning~\cite{bourtoule2021machine} provides a formal framework for selective removal of data, concepts, or behaviours from trained models. In diffusion models, most prior work has focused on either \emph{concept unlearning}~\cite{gandikota2023erasing, gandikota2024unified, kumari2023ablating}, which suppresses high-level semantics like artistic styles or unsafe content, or on \emph{data unlearning}~\cite{alberti2025data, heng2023selective, wu2025erasediff}, which aims to remove specific training images. Identity unlearning is related but distinct: in ID-conditioned models, identity is not controlled by discrete text prompts, but rather by continuous, structured feature representations, making selective forgetting substantially more delicate.
Consequently, identity unlearning has only recently emerged as a distinct research direction. While GUIDE~\cite{seo2024generative}, LEGATO~\cite{chen2026legato}, and Forget-Me~\cite{qi2025forget} study identity forgetting in different settings, the closest prior works exploring identity erasure in diffusion models include BIA~\cite{shaheryar2025black} and WID~\cite{shaheryar2025unlearn}. However, BIA targets standard diffusion probabilistic models rather than ID-conditioned models like Arc2Face~\cite{papantoniou2024arc2face}, while WID relies on a custom, unreleased base model, limiting its reproducibility. Thus, a critical gap remains between the conceptual formulation of identity unlearning in ID-conditioned diffusion models and its practical realization.

To address this gap, we present \textbf{P}roximity-guided \textbf{I}dentity \textbf{U}nlearning (\textbf{PIU}), a framework designed for surgical identity erasure in ID-conditioned diffusion models like Arc2Face~\cite{papantoniou2024arc2face}. Unlike methods that attempt to naively suppress a conditioning signal, PIU formulates forgetting as an explicit replacement and reassigns the target identity to a safe anchor identity within the learned identity space, as shown in Figure~\ref{fig:manual_identity_grid}. To achieve this, PIU combines a proximity-based anchor selection strategy, grounded in the geometry of ArcFace~\cite{deng2019arcface} identity embeddings, with an anchor-guided unlearning objective tailored for diffusion architectures. Furthermore, we provide an empirical analysis of identity-sensitive layers in Arc2Face, enabling a more surgical and efficient fine-tuning procedure that maximizes identity erasure while minimizing negative impact on the remaining identity manifold. 
Extensive experiments on the state-of-the-art Arc2Face model~\cite{papantoniou2024arc2face} demonstrate that PIU effectively removes target identities from CelebA-HQ~\cite{karras2018progressive} while preserving the quality and identity consistency of images for retained, non-target subjects.  
Through a reproducible benchmark, both quantitative and qualitative results show that identity unlearning in ID-conditioned models is feasible and distinct from prior concept or data unlearning paradigms. Our contributions are summarized as follows:
\begin{itemize}
    \item We propose PIU, to our knowledge the first reproducible framework for identity unlearning in ID-conditioned diffusion models, namely Arc2Face~\cite{papantoniou2024arc2face}.
    %\item We propose an anchor-guided replacement objective that removes the target identity by reassigning it to a selected anchor identity in the learned identity space. 
    \item We formulate unlearning as explicit geometric replacement, erasing the target identity by reassigning it to an anchor selected via a novel proximity-based strategy within the continuous identity space.
    %\item We introduce a proximity-based strategy for anchor identity selection, motivated by the geometry of ArcFace representations.
    \item We provide a layer-level analysis of identity-sensitive components in Arc2Face, enabling surgical and efficient fine-tuning that targets the most relevant weights.
    %\item We achieve stronger quantitative and qualitative results than comparable baselines for identity unlearning.
    \item We compare PIU to the state-of-the-art through extensive experiments and demonstrate that it delivers effective identity unlearning while avoiding degradation for non-target identities observed in prior works.
\end{itemize}

\section{Related work}
\label{sec:RelatedWork}

\paragraph{Image generation.}
Image synthesis has advanced rapidly with the development of deep generative models, progressing from Generative Adversarial Networks~(GANs)~\cite{goodfellow2014generative, karras2017progressive, karras2019style} to diffusion models~\cite{Ho2020DDPM}. In particular, Latent Diffusion Models (LDMs)~\cite{rombach2022high} have nowadays established a robust foundation for efficient and high-quality conditional synthesis by operating in a lower-dimensional latent space. Building on this framework, many works addressed identity-driven face generation. Early methods personalized generation from a few subject images through embedding optimization~\cite{gal2022image} and model fine-tuning~\cite{ruiz2023dreambooth}, with recent advancements incorporating identity-based training objectives to enhance identity consistency~\cite{tomavsevic2025id}. To bypass the need for per-subject optimization, other approaches introduced adapter modules~\cite{ye2023ip, li2024photomaker, xiao2025fastcomposer}, which extract subject-aware visual features using pretrained image encoders and inject them via cross-attention layers. In contrast, Arc2Face~\cite{papantoniou2024arc2face} enables identity-consistent generation by injecting identity embeddings from a pretrained recognition model~\cite{deng2019arcface} into the text encoder by replacing token embeddings of a template prompt. This approach achieves exceptional identity preservation, establishing Arc2Face as a core framework for generating high-quality, diverse, and identity-consistent face images suitable for training recognition models.

\vspace{-2mm}

\begin{figure*}[t]
  \centering
  \includegraphics[trim={5mm 0 12mm 7mm},clip, width=0.9\linewidth]{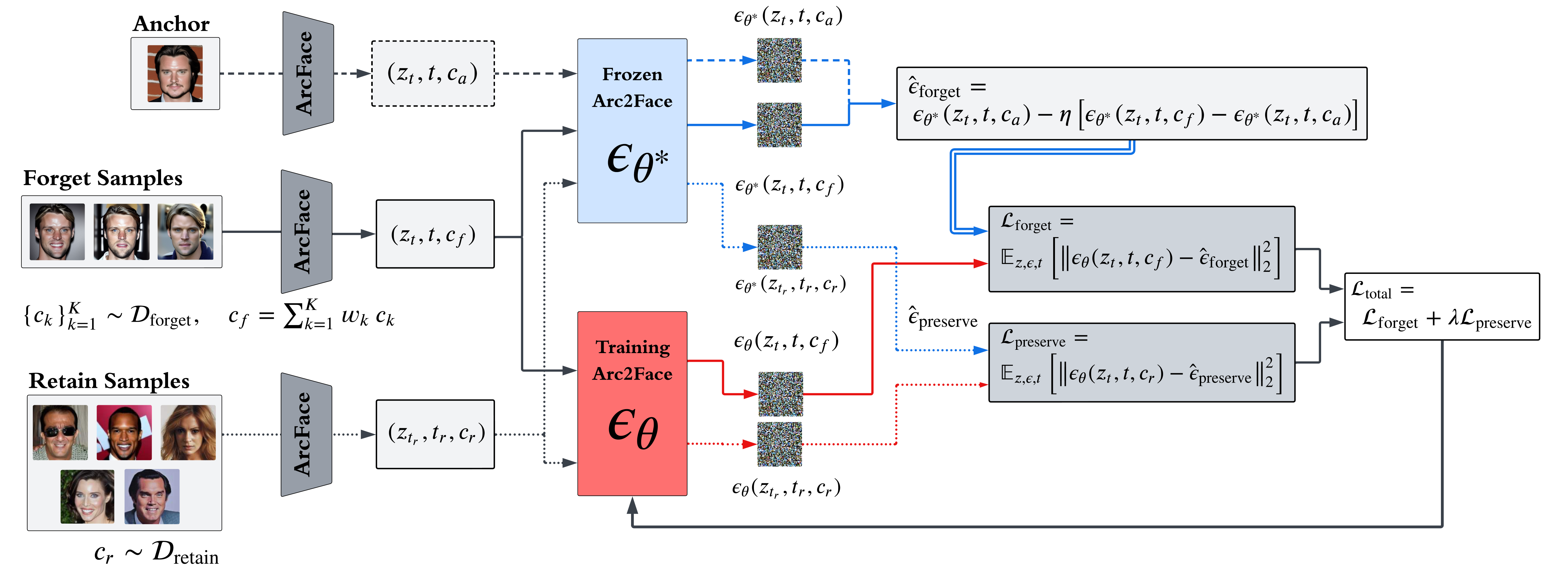}
  \caption{\textbf{Overview of the proposed Proximity-guided Identity Unlearning (PIU) framework.} During training, target (forget) and retain identity embeddings are processed by frozen and trainable versions of Arc2Face~\cite{papantoniou2024arc2face}. The forget loss $\mathcal{L}_{\mathrm{forget}}$ steers the trainable prediction toward a proximity-based anchor trajectory, while the preservation loss $\mathcal{L}_{\mathrm{preserve}}$ enforces alignment with the frozen model for non-target identities. The joint objective is optimized to ensure efficient identity removal while preserving the generative prior of the model.
  }
  \label{fig:method}
  \vspace{-4mm}
\end{figure*}

\paragraph{Identity unlearning.} Modern diffusion models have raised privacy and copyright concerns due to their capacity to reproduce training samples~\cite{somepalli2023diffusion,carlini2023extracting,somepalli2023understanding}. This memorization risk, coupled with the requirements of legal frameworks such as the General Data Protection Regulation (GDPR)~\cite{protection2018general}, has motivated the development of machine unlearning techniques.  The most widely explored direction is \textit{concept unlearning} for text-to-image models. Early iterative fine-tuning methods like Erased Stable Diffusion (ESD)~\cite{gandikota2023erasing} suppress concepts by steering the conditional noise prediction toward its unconditional counterpart. Subsequent works explore minimalist unlearning through sparse weight masking~\cite{zhang2025minimalist}, while closed-form approaches like Unified Concept Editing (UCE)~\cite{gandikota2024unified} compute analytic updates to targeted model weights, typically in cross-attention layers.
Another prominent direction is \textit{data unlearning}, which aims to remove specific training samples so they cannot be reproduced~\cite{fan2023salun, heng2023selective}. Within this paradigm, Subtracted Importance Sampled Scores (SISS)~\cite{alberti2025data} introduces an importance-sampled loss function providing theoretical guarantees and empirical evidence for erasing specific identities.
Despite this progress, \textit{identity unlearning} remains largely underexplored, especially for identity-conditioned diffusion models. 
Unlike standard text-to-image models where concepts are tied to discrete text tokens, these architectures condition image generation on dense identity embeddings~\cite{papantoniou2024arc2face}, making unlearning highly sensitive to how the target is redirected within the identity space.
While GUIDE~\cite{seo2024generative} introduced anchor-based identity unlearning, the mechanisms remained limited to GAN-based generators. For diffusion models, Forget-Me~\cite{qi2025forget} studies identity forgetting in a federated setting, but its reliance on decentralized client updates renders it unsuitable for centralized models like Arc2Face~\cite{papantoniou2024arc2face}. 
The closest prior works are Black Hole-Driven Identity Absorbing (BIA)~\cite{shaheryar2025black} and the Unlearn and Protect variant With ID-guidance (WID)~\cite{shaheryar2025unlearn}. 
While BIA~\cite{shaheryar2025black} transfers anchor-guided unlearning to the $h$-space of standard diffusion models, it is not designed for ID-conditioned architectures and lacks publicly available code. 
To our knowledge, 
WID~\cite{shaheryar2025unlearn} is the only prior work explicitly addressing identity unlearning in ID-conditioned diffusion models. However, it relies on a simple identity-guided objective lacking both a dedicated preservation strategy to protect non-target identities and a principled mechanism for anchor selection. Furthermore, because it utilizes a custom, unreleased base model, direct reproduction is severely hindered. 
Our method addresses these gaps by introducing an explicit preservation term and a proximity-based anchor selection strategy on top of a publicly available framework, establishing a robust and reproducible identity unlearning pipeline.

% \begin{figure*}[t]
%   \centering
%   \includegraphics[trim={5mm 0 12mm 7mm},clip, width=\linewidth]{imgs/diagram.png}
%   \caption{\textbf{Overview of the proposed Proximity-guided Identity Unlearning (PIU) framework.} During training, target (forget) and retain identity embeddings are processed by frozen and trainable versions of Arc2Face~\cite{papantoniou2024arc2face}. The forget loss $\mathcal{L}_{\mathrm{forget}}$ steers the trainable prediction toward a proximity-based anchor trajectory, while the preservation loss $\mathcal{L}_{\mathrm{preserve}}$ enforces alignment with the frozen model for non-target identities. The joint objective is optimized to ensure efficient identity removal while preserving the generative prior of the model.
%   }
%   \label{fig:method}
%   \vspace{-4mm}
% \end{figure*}

\section{Methodology}
 
% This section presents our targeted identity unlearning approach for the Arc2Face identity-conditioned diffusion model, depicted in Figure \ref{fig:method}. Following the preliminaries (Section \ref{subsec:preliminaries}), we introduce Anchor Guided Identity Unlearning (Section \ref{subsec:anchor_guided_unlearning}), which steers the target identity toward a selected anchor rather than an unconditional prior. To maximize unlearning efficacy while preserving generative capabilities of the model, we propose a proximity-based anchor selection strategy in the ArcFace identity space (Section \ref{subsec:threshold_optimized_anchor_selection}). Finally, Section \ref{subsec:surgical_layers} describes our condition-driven training strategy and the surgical layer selection used to minimize network disruption.

This section presents our Proximity-guided Identity Unlearning (PIU) approach for the identity-conditioned Arc2Face~\cite{papantoniou2024arc2face} diffusion model, which steers the target identity toward a proximity-based anchor rather than an unconditional prior, as depicted in Figure~\ref{fig:method}. % via condition-driven fine-tuning of surgical layers to minimize network disruption. 

\subsection{Identity-conditioned Latent Diffusion}
\label{subsec:preliminaries}

Latent Diffusion Models (LDMs)~\cite{rombach2022high} improve the efficiency of standard denoising diffusion probabilistic models~\cite{Ho2020DDPM} by operating in a compressed latent space of a pretrained variational autoencoder. Given an image $x_0$, the encoder first maps it to a lower dimensional latent representation $z_0 = \mathcal{E}(x_0)$. The forward diffusion process then gradually corrupts $z_0$ over timesteps $t \in \{0, \dots, T\}$ by adding Gaussian noise according to a predefined schedule, so that $z_t = \sqrt{\bar{\alpha}_t}\, z_0 + \sqrt{1 - \bar{\alpha}_t}\, \epsilon$, where $\epsilon \sim \mathcal{N}(0, I)$. The reverse process can be learned by a denoising network, typically a U-Net~\cite{ronneberger2015u}, that is trained to predict the added noise $\epsilon$ at each timestep $t$ as $\epsilon_\theta(z_t, t, c)$ with $c$ as the conditioning information. The denoised sample $\hat{z}_0$ can then be mapped back to the image space through the decoder as $\hat{x}_0 = \mathcal{D}(\hat{z}_0)$. At inference time, conditional generation can even be strengthened with classifier-free guidance (CFG)~\cite{ho2022classifier}, which combines unconditional and conditional predictions. In text-to-image implementations of LDMs like Stable Diffusion~\cite{rombach2022high}, the condition $c$ is derived from textual prompts processed by a CLIP text encoder~\cite{radford2021learning}.

Arc2Face~\cite{papantoniou2024arc2face} adapts this text-conditioned framework for identity-driven generation of face images. Rather than relying on open text prompts, the conditioning space is restricted to facial recognition features. To this end, Arc2Face extracts an identity embedding $v \in \mathbb{R}^{512}$ with a pretrained ArcFace~\cite{deng2019arcface} recognition network. This vector is zero-padded to $\hat{v} \in \mathbb{R}^{768}$ and substituted for the \texttt{<id>} token within a template prompt \textit{``photo of a \texttt{<id>} person''}. The CLIP~\cite{radford2021learning} text encoder then processes this sequence to produce the final identity-encoded condition $c_{\mathrm{id}}$. Arc2Face trains the denoising network to predict $\epsilon_\theta(z_t, t, c_{\mathrm{id}})$, with $c_{\mathrm{id}}$ encoding the target identity, thus replacing the flexibility of textual conditioning with identity conditioning.

\subsection{Anchor-Guided Identity Unlearning}
\label{subsec:anchor_guided_unlearning}
The goal of PIU is to erase a target identity from an identity-conditioned diffusion model, in our case Arc2Face~\cite{papantoniou2024arc2face}. Methods for prompt-based concept unlearning are not directly applicable in this setting, due to the fixed prompt constraints of the architecture~\cite{papantoniou2024arc2face}. For this reason, we build on the core ideas of concept unlearning, namely Erased Stable Diffusion (ESD)~\cite{gandikota2023erasing}, and fundamentally extend this paradigm for geometrically-informed unlearning tailored to the specific structure of the identity space. 

Let $\theta^*$ denote the parameters of the frozen original model, and $\theta$ be the parameters of the trainable model to be unlearned. Given a noisy latent $z_t = \mathcal{E}(x_t)$ at timestep $t$, let $c_f$ be the condition embedding of the forget identity. In standard ESD~\cite{gandikota2023erasing}, concept unlearning is achieved by pushing the conditional prediction away from the target concept and toward the unconditional prior $\varnothing$. In our setting, however, steering the forget identity toward the unconditional prediction is suboptimal, since the unconditional prior still maps to a generic identity.
Consequently, using it as a target can unpredictably entangle identity unlearning with non-identity features, leading to attribute changes or even severe quality degradation.
Instead, we propose to steer the forget identity $c_f$ toward an anchor identity condition $c_a$. We define the prediction for the forget target as:
{\small\begin{equation}
\hat{\epsilon}_{\mathrm{forget}} = \epsilon_{\theta^*}(z_t,t,c_a) 
- \eta \left[ \epsilon_{\theta^*}(z_t,t,c_f) - \epsilon_{\theta^*}(z_t,t,c_a) \right],
\label{eq:epsilon_forget}
\end{equation}
}
where $\eta$ is the negative guidance scale. The forget loss then forces the trainable model, conditioned on the forget identity, to match this modified trajectory:
\begin{equation}
\mathcal{L}_{\mathrm{forget}} = \mathbb{E}_{z,\epsilon,t} \left[ \left\| \epsilon_\theta(z_t,t,c_f) 
- \hat{\epsilon}_{\mathrm{forget}} \right\|_2^2 \right].
\end{equation}

Since identities are highly interconnected within the U-Net~\cite{ronneberger2015u} weights, solely optimizing for the forget objective may degrade the generation of non-target identities. To alleviate this, we introduce an identity preservation loss, inspired by recent fine-tuning methods~\cite{ruiz2023dreambooth}. For retain identities $c_r$, we enforce the trainable model to remain aligned with the predictions of the frozen model $\hat{\epsilon}_{\mathrm{preserve}} = \epsilon_{\theta^*}(z_t,t,c_r)$ through the identity preservation objective as:
\begin{equation}
\mathcal{L}_{\mathrm{preserve}} = \mathbb{E}_{z,\epsilon,t} \left[ \left\| \epsilon_\theta(z_t,t,c_r)
- \hat{\epsilon}_{\mathrm{preserve}}  \right\|_2^2 \right].
\end{equation}
The final training objective is then the weighted sum:
\begin{equation}
\mathcal{L}_{\mathrm{total}} = \mathcal{L}_{\mathrm{forget}} + \lambda \mathcal{L}_{\mathrm{preserve}},
\end{equation}
where the preservation weight $\lambda$ balances forgetting and model preservation: if too large, unlearning weakens; if too small, non-target identities may drift. We use $\lambda=10$ by default and defer its analysis to Section~\ref{subsec:ablations}.
Figure \ref{fig:method} depicts this process. PIU suppresses identity-specific information associated with $c_f$ by redirecting it toward the behavior induced by the anchor condition $c_a$, while retaining the rest of the model knowledge through the preservation term.

\subsection{Proximity-based Anchor Selection}
\label{subsec:threshold_optimized_anchor_selection}

{%\color{blue}
The most critical mechanism in our PIU method is the selection of the anchor identity $c_a$. Arc2Face relies on ArcFace embeddings~\cite{deng2019arcface} as the identity representation, which are learned using a margin based objective designed to reduce intra class variability while maximizing inter class separation~\cite{schroff2015facenet}. This induces a hyperspherical geometry and makes the embedding space particularly suitable for reasoning about identity similarity, where the intrinsic distance between identities is defined by the geodesic distance $d_{\mathbb{S}}(\mu_f, \mu_a) = \arccos(\langle \mu_f, \mu_a \rangle)$.
However, for unlearning in this space, anchor selection must strike a careful balance. If the anchor identity $c_a$ is too similar to the forget identity $c_f$, it yields near-zero gradient updates. 
Assuming local smoothness, the conditional displacement satisfies $\epsilon_{\theta^*}(z_t,t,c_f) - \epsilon_{\theta^*}(z_t,t,c_a) \approx J_\epsilon(z_t, t, c_f)(c_f - c_a) \approx 0$, resulting in weak unlearning.
Conversely, if the anchor is placed too far across this manifold (the antipodal regime where $d_{\mathbb{S}} \approx \pi$), steering the generation across empty regions may distort the generative prior that the preservation loss cannot mitigate. Therefore, optimal anchors should lie at an intermediate geodesic distance, sufficiently separated to induce identity removal but close enough to preserve the facial prior.
}

To resolve this, we formalize a threshold guided selection strategy. Let $\mathcal{D}$ be the training dataset partitioned by identity. For each image $x \in \mathcal{D}$, we extract its ArcFace embedding $w(x) \in \mathbb{R}^{512}$. Then, for each identity $i$, we compute a robust and representative centroid of the identity
\begin{equation}
\label{eq::centroid}
\mu_i = \frac{1}{|\mathcal{D}_i|} \sum_{x \in \mathcal{D}_i} w(x),
\end{equation}
where $\mathcal{D}_i$ denotes the images of identity $i$. 
% This centroid provides a robust identity-level representation, which is especially meaningful given the robustness of the sub-center ArcFace~\cite{deng2019arcface} embeddings to noisy intra-class samples.
Let $\mu_f$ be the centroid of the forget identity. For every candidate identity $j \neq f$, we compute the cosine similarity $s_j = \frac{\langle \mu_f, \mu_j \rangle}{\|\mu_f\|_2 \|\mu_j\|_2}$. Then, instead of selecting the most similar or dissimilar identity, we define the candidate anchor set $\mathcal{A}_\tau = \left\{ j \neq f \;:\; \left| s_j - \tau \right| < \varepsilon \right\}$,
with tolerance $\varepsilon = 10^{-2}$, and uniformly sample the anchor identity $a \sim \mathcal{A}_\tau$. Empirically, we find that $\tau = 0.2$ provides the best trade-off in ArcFace~\cite{deng2019arcface} space, favoring an anchor that is sufficiently close to the forget identity to preserve a realistic facial prior, yet sufficiently separated to effectively redirect the generations away from the target identity.

\subsection{Condition-driven and Localized fine-tuning}
\label{subsec:surgical_layers}

% \begin{figure}[t]
%     \centering
%     \begin{subfigure}{\linewidth}
%         \centering
%         \includegraphics[width=\linewidth]{imgs/layer_average_comparison.png}
%         \caption{Average intra-class and inter-class cosine similarity of cross-attention KV activations across U-Net layers.}
%         \label{fig:CA-KV}
%     \end{subfigure}
    
%     \vspace{0.5em}
    
%     \begin{subfigure}{\linewidth}
%         \centering
%         \includegraphics[width=\linewidth]{imgs/layer_query_interclass_comparison.png}
%         \caption{Average inter-class cosine similarity of cross-attention Q activations across U-Net layers for different identity-similarity thresholds.}
%         \label{fig:Q-activations}
%     \end{subfigure}
    
%     \caption{Layer-wise cosine similarity analysis of cross-attention activations in Arc2Face. The most identity-sensitive behavior is concentrated in the third downsampling block, the middle block, and the first two upsampling blocks. \TODO{(a) for KV, (b) for Q, could we have only one?}}
%     \label{fig:cross_attention_analysis}
%     \vspace{-4mm}
% \end{figure}

\begin{figure}[t]
    \centering
    \includegraphics[width=\linewidth]{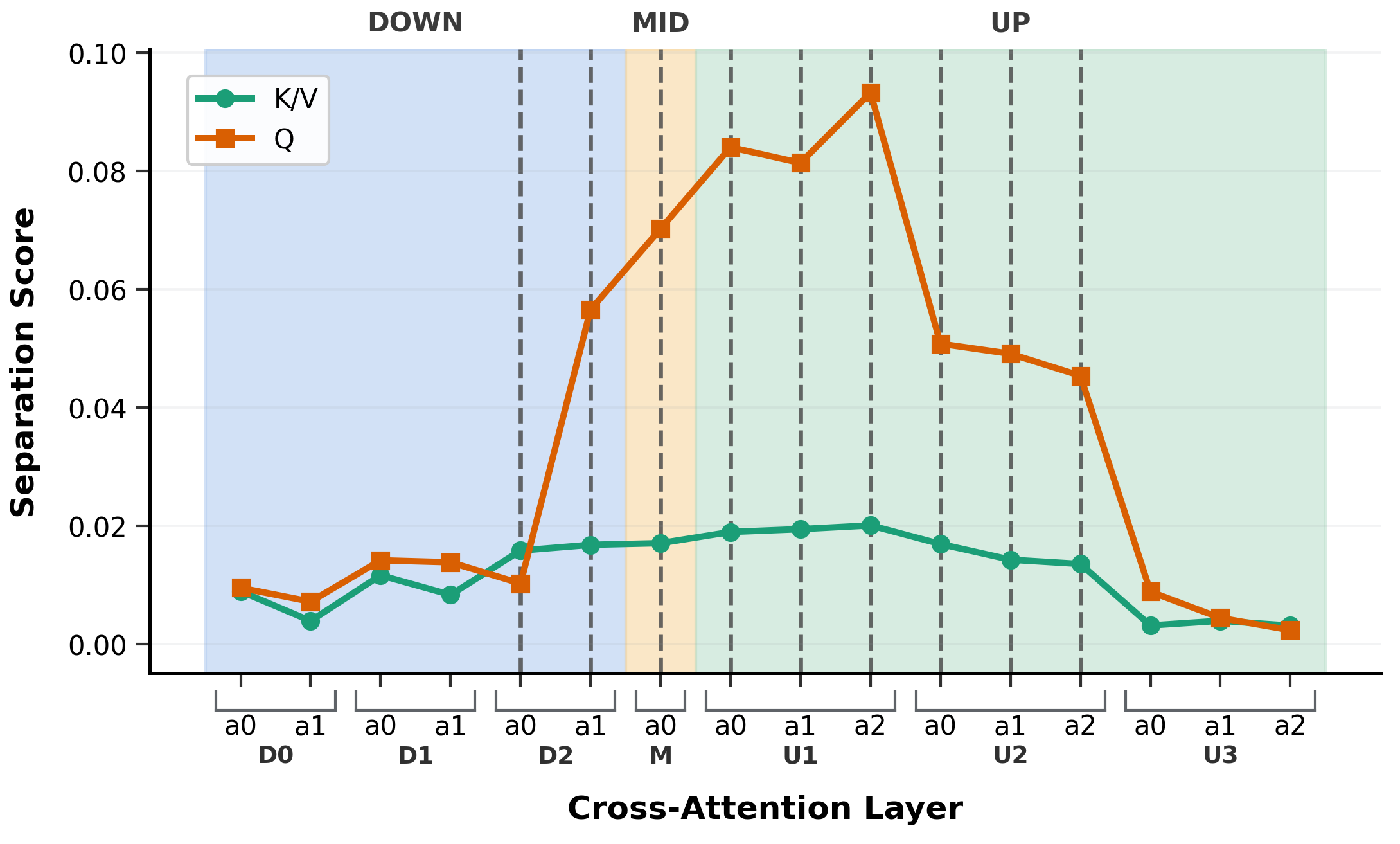}
    \caption{\textbf{Layerwise identity separation in Arc2Face cross-attention activations.} Identity separation captures the ability of the model to distinguish between identities across Keys (K), Values (V), and Queries (Q). Higher scores indicate stronger identity conditioning. Dashed lines denote the selected surgical blocks: Downsampling (D2), Middle (M), and Upsampling (U1 \& U2).
    }
    \label{fig:cross_attention_analysis}
    \vspace{-6mm}
\end{figure}
% \TODO{We define $S^{KV}_{\ell}=\mathrm{Intra}^{KV}_{\ell}-\mathrm{Inter}^{KV}_{\ell}$ and $S^{Q}_{\ell}=1-\frac{1}{|T|}\sum_{t\in T}\mathrm{Inter}^{Q}_{\ell,t}$. $\mathrm{Intra}$ and $\mathrm{Inter}$ denote mean cosine similarity within the same identity and across different identities, respectively. We use sensitive denoising timesteps $T={750,1000}$ where query features change most for different IDs. K/V activations are much less  $t$-conditioned than $Q$, so we treat them as approximately timestep-stable and compute KV separation at a single representative $t$. Higher scores mean stronger identity separation in XAttn activations, indicating layers where the network attends more to identity conditioning. Vertical dashed lines mark the surgical XAttn blocks used for unlearning (\texttt{down\_blocks.2}, \texttt{mid\_block}, \texttt{up\_blocks.1}, \texttt{up\_blocks.2}).}

The main object of unlearning in our generative setting is the identity condition itself rather than a specific training image. Accordingly, we do not rely on latents encoded from the original training images. Instead, the latent input is sampled directly from Gaussian noise, $z_0 \sim \mathcal{N}(0,I)$, so that the optimization focuses primarily on the conditioning signal rather than on the content of specific real images.

To avoid restricting unlearning to a finite set of observed forget embeddings, we further extend the forget set by sampling synthetic identity conditions as convex combinations of forget-identity embeddings. Concretely, let $\{c_k\}_{k=1}^K \sim \mathcal{D}_{\mathrm{forget}}$ be identity embeddings associated with the forget identity, and sample mixing weights $w \sim \mathrm{Dir}(\alpha)$. We then define the forget condition as $c_f = \sum_{k=1}^{K} w_k c_k$, where $\sum_{k=1}^{K} w_k = 1$ and $w_k \geq 0$. This way, the model is encouraged to forget the local identity region of the target subject, rather than only a finite set of observed embeddings.

Which parameters to fine-tune represents another critical design decision that impacts both computational efficiency and unlearning efficacy. To this end, prior methods focus on cross-attention weights~\cite{gandikota2023erasing}, which are commonly utilized for conditioning, while other approaches use gradient-based saliency or empirical activations to determine concept-specific weights~\cite{fan2023salun,liu2024implicit,wu2025erasediff}. Building on these insights, we measure layerwise identity separation within cross-attention blocks of the Arc2Face U-Net~\cite{ronneberger2015u}, across key, value, and query projections. 
Specifically, we define Key and Value (K/V) separation for layer $\ell$ as the gap between mean intra- and inter-identity cosine similarity, i.e., $S^{KV}_{\ell} = \mathrm{Intra}^{KV}_{\ell} - \mathrm{Inter}^{KV}_{\ell}$. Because K/V activations are projected directly from the global identity condition, they are inherently timestep-independent, allowing evaluation at a single representative timestep. Conversely, Query (Q) features are highly timestep-dependent and sensitive to initial noise. Thus, we compute Q separation based solely on inter-identity divergence as $S^{Q}_{\ell} = 1 - \frac{1}{|T|}\sum_{t\in T}\mathrm{Inter}^{Q}_{\ell,t}$, utilizing  sensitive denoising timesteps $T=\{750, 1000\}$ where query features vary most across identities.

As shown in Figure~\ref{fig:cross_attention_analysis}, the strongest separation concentrates in the third downsampling block, the middle block, and the first two upsampling blocks within the cross-attention modules. We refer to these as the \emph{surgical layers}. 
% Restricting fine-tuning to this subset modifies only a small fraction of the total model parameters, approximately $4.29\%$, making the method notably more efficient while retaining strong unlearning performance.
Restricting fine-tuning to this subset modifies only a small fraction of the total model parameters ($4.29\%$), making the method more efficient while achieving superior selective unlearning by targeting identity-sensitive components.

% Parameter selection constitutes a critical design decision, influencing both computational efficiency and unlearning efficacy. While foundational approaches~\cite{gandikota2023erasing} focus exclusively on cross attention weights as the primary carriers of conditioning information, more recent strategies~\cite{fan2023salun,liu2024implicit,wu2025erasediff} utilize gradient based saliency or empirical activations to identify influential parameters. Building on these insights, we analyze the activation patterns within the Arc2Face U-Net to identify a highly localized subset of weights responsible for identity encoding. We designate these as surgical layers. By restricting fine tuning to this minimal subset, we achieve robust identity erasure while preserving the integrity of the broader generative prior.

\section{Experiments and Results}
\label{sec:experiments}

%\subsection{Experimental Setup}
%\label{subsec:experimental_setup}
\paragraph{Implementation details.}
We perform unlearning on Arc2Face~\cite{papantoniou2024arc2face}, an identity-conditioned latent diffusion model built on the Stable Diffusion 1.5~\cite{rombach2022high}. Following their configuration, images are generated at a resolution of $512\times512$ using the multistep DPM-Solver++\cite{lu2025dpm} with $25$ denoising steps and a classifier-free guidance scale of $3$.
Ablation experiments are run first for $300$ steps with a batch size of $32$, using AdamW with learning rate $10^{-4}$, $\beta_1=0.9$, $\beta_2=0.999$, $\epsilon=10^{-8}$, and weight decay $0.01$, along with default unlearning hyperparameters $\lambda=10$, $\tau=0.2$, and $\eta=1.5$. At each training step, the loss is computed from $32$ forget and $32$ retain identity embeddings extracted with the ArcFace~\cite{deng2019arcface} recognition model. For the comparison with the state-of-the-art, unlearning is run for $400$ steps with the best performing parameters from the ablation studies.

\vspace{-4mm}
\paragraph{Datasets.} We conduct experiments on CelebA-HQ~\cite{karras2018progressive}, a high-quality face dataset used in the training of Arc2Face~\cite{papantoniou2024arc2face}, to ensure 
{%\color{blue} 
that measured performance drops reflect true identity erasure rather than failures caused by out-of-distribution generations.} The dataset contains $30$k images at resolution $512\times512$, split into $28$k training and $2$k validation images; we use only the training split. Since CelebA-HQ does not provide the identity partition required for our evaluation, we extract identity embeddings with ArcFace~\cite{deng2019arcface} for each image and cluster the embeddings with DBSCAN~\cite{ester1996density} in cosine space, using $\epsilon=0.35$ and a minimum cluster size of two images. %$\texttt{min\_samples}=3$. 
This yields $27{,}996$ successfully processed images grouped into $9{,}683$ identity clusters. Details are deferred to the supplementary material.

Each unlearning experiment targets a single forget identity, splitting its images into forget training and validation subsets ($65/35$ ratio). The remaining images form the retain set, with a $90/10$ split between training and unseen retain evaluation. During unlearning, the forget training set is extended as described in Section~\ref{subsec:surgical_layers} to generate conditioning samples up to the batch size. Each training step additionally samples a batch of retain conditions from the retain training pool, while the anchor is set as the centroid of the anchor identity. 
{%\color{blue}
For the comparison with the state-of-the-art, we repeat this procedure independently over $50$ forget identities, always starting from the original pretrained Arc2Face model. To prevent the demographic skew inherent to random sampling, these identities are manually curated to ensure diverse representation across race and gender with with an average cluster size of $15$ images per identity, while retaining variance in pose, expression, and lighting, as shown in the supplementary material.}
Evaluation uses $25$ generated images for each of the forget and retain test splits.

\vspace{-4mm}
\paragraph{Evaluation metrics.}
We assess identity removal using \emph{Identity Score Matching} (ISM)~\cite{van2023anti}, which separately measures the identity similarity of generated forget and retain samples to their identity centroid in the space of a pretrained ArcFace recognition model~\cite{deng2019arcface}. 
Since centroid similarity alone does not fully characterize forgetting, we also report the \emph{Selective Removal and Keep} (SRK)~\cite{shaheryar2025unlearn} measure, which jointly evaluates forgetting and preservation via nearest-centroid classification, providing a stronger notion of selective identity removal. {%\color{blue} 
Details are provided in the supplementary material, along with additional results using the AdaFace model\cite{kim2022adaface}, to mitigate potential circularity from evaluating in the conditioning feature space.}

% To mitigate potential circularity from evaluating in the same feature space used for conditioning, we further validate cross-space unlearning transfer by computing ISM and SRK using the AdaFace recognition model, with full results detailed in the supplementary material (Section~\ref{sec:supp_unlearning_metrics}).

% and \emph{Selective Removal and Keep} (SRK)~\cite{shaheryar2025unlearn}; formal definitions and implementation details are deferred to Section~\ref{sec:supp_unlearning_metrics}. ISM measures the similarity of generated samples to the corresponding identity centroid in ArcFace space, and is reported for both forget and retain identities. Since centroid similarity alone does not fully characterize successful forgetting, we also report SRK, which jointly evaluates forgetting and preservation through nearest-centroid classification, providing a stronger notion of selective identity removal.

For image quality and realism, we rely on a \emph{Kernel-based distribution Distance} (KD)~\cite{gretton2006kernel} in the feature space of DINOv2~\cite{oquab2024dinov2}, implemented as an unbiased Maximum Mean Discrepancy estimator~\cite{stein2023exposing}. Compared to Fréchet Inception Distance~\cite{heusel2017_FID}, this measure better suits small-sample regimes and captures finer facial properties. We report it as $\Delta$KD, computed between the retain validation distribution and generated forget or retain samples after unlearning, where lower values indicate closer alignment with the reference distribution. We complement this with the \emph{eDIFFIQA}~\cite{babnik2024ediffiqa} quality measure, which estimates the suitability of face images for recognition. Both $\Delta$KD and eDIFFIQA averaged over all evaluated identities.

\subsection{Comparison with the State-of-the-art}

To evaluate the effectiveness of our PIU method, we benchmark its performance against representative state-of-the-art methods from distinct unlearning paradigms, specifically SISS~\cite{alberti2025data} for data unlearning, UCE~\cite{gandikota2024unified} for concept unlearning, and WID~\cite{shaheryar2025unlearn}, to our knowledge the only prior method explicitly designed for identity unlearning in an ID-conditioned diffusion model. To ensure a fair comparison, we use identical identities, seeds, and anchors for all methods. 
Details regarding the adaptation of these methods for Arc2Face~\cite{papantoniou2024arc2face} are provided in the supplementary material.

% \TODO{   The final method evaluations in Table~\ref{tab:unlearning_results} show that PIU outperforms all other methods across all unlearning metrics.}

% The final method evaluations in Table~\ref{tab:unlearning_results} show that PIU outperforms all other methods across all unlearning metrics. Identical identities, seeds, and anchors (where applicable) were used to ensure a fair comparison. As baselines, we reproduce SISS, a strong recent method for data unlearning (Section~\ref{subsec:siss_reproduction}), UCE, a representative concept-unlearning approach (Section~\ref{subsec:uce_reproduction}), and WID, to the best of our knowledge the only prior method explicitly designed for identity unlearning in an ID-conditioned diffusion model (Section~\ref{subsec:wid_reproduction}).

% To evaluate the efficacy of PIU, we conduct a comparative analysis against representative state of the art methods across three unlearning paradigms. We include SISS~\cite{...} as a baseline for data unlearning, UCE~\cite{...} for general concept erasure, and WID~\cite{...}, which represents the only prior framework specifically targeting identity unlearning in identity conditioned diffusion models. For a rigorous comparison, all methods utilize identical identities, random seeds, and anchor configurations where applicable. As summarized in Table~\ref{tab:unlearning_results}, PIU consistently outperforms these baselines across all established unlearning metrics.

\vspace{-2mm}
\paragraph{Quantitative results.}
\label{subsec:quantitative_results} 
Results in Table~\ref{tab:unlearning_results} demonstrate that PIU outperforms all prior methods across all unlearning and utility metrics. SISS and UCE perform poorly for unlearning in ID-conditioned models. WID achieves comparable forget ISM to PIU, but (i) requires an encoder inference pass during each training step, and (ii) degrades the general model, as evidenced by the drop in retain ISM. Consequently, PIU achieves the highest SRK result.

\begin{table}[!t]
\caption{\textbf{Comparison with other unlearning methods.} PIU achieves the strongest identity unlearning (lowest forget ISM and highest SRK) while preserving non-target identities and image quality (best retain ISM, $\Delta$KD, and eDIFFIQA scores). Bold denotes the best result for each metric; underline the second best.}
\label{tab:unlearning_results}
\centering
%\footnotesize
\setlength{\tabcolsep}{3pt}
\resizebox{\columnwidth}{!}{%
\begin{tabular}{l|cc|c|cc|cc}
\hline 
 & \multicolumn{2}{c|}{\textbf{ISM}} & \textbf{SRK} $\uparrow$ & \multicolumn{2}{c|}{\textbf{$\Delta$KD}} & \multicolumn{2}{c}{\textbf{eDIFFIQA}} \\
\cline{2-3} \cline{4-4} \cline{5-6} \cline{7-8}
\textbf{Model} & Forget $\downarrow$ & Retain $\uparrow$ & & Forget $\downarrow$ & Retain $\downarrow$ & Forget $\uparrow$ & Retain $\uparrow$ \\
\hline
Arc2Face \cite{papantoniou2024arc2face} & 0.78 & 0.75 & 0.99 & -- & -- & 0.76 & 0.76 \\
\hline
SISS \cite{alberti2025data}    & 0.59 & 0.58 & 0.98 & 5.23 & 8.14 & 0.65 & \underline{0.64} \\
UCE  \cite{gandikota2024unified}  & 0.59 & \underline{0.70} & 1.00 & \textbf{3.69} & \underline{1.26} & \underline{0.75} & \textbf{0.75} \\
WID \cite{shaheryar2025unlearn}     & \underline{0.32} & 0.44 & \underline{49.28} & \underline{4.35} & 3.36 & \textbf{0.76} & \textbf{0.75} \\
\textbf{PIU} (Ours)& \textbf{0.31} &\textbf{0.72} & \textbf{88.96} & 8.12 & \textbf{0.39} & \underline{0.75} & \textbf{0.75} \\
\hline
\end{tabular}
}
\vspace{-2mm}
\end{table}

\begin{figure}[t]
    \centering
    \includegraphics[width=0.85\linewidth]{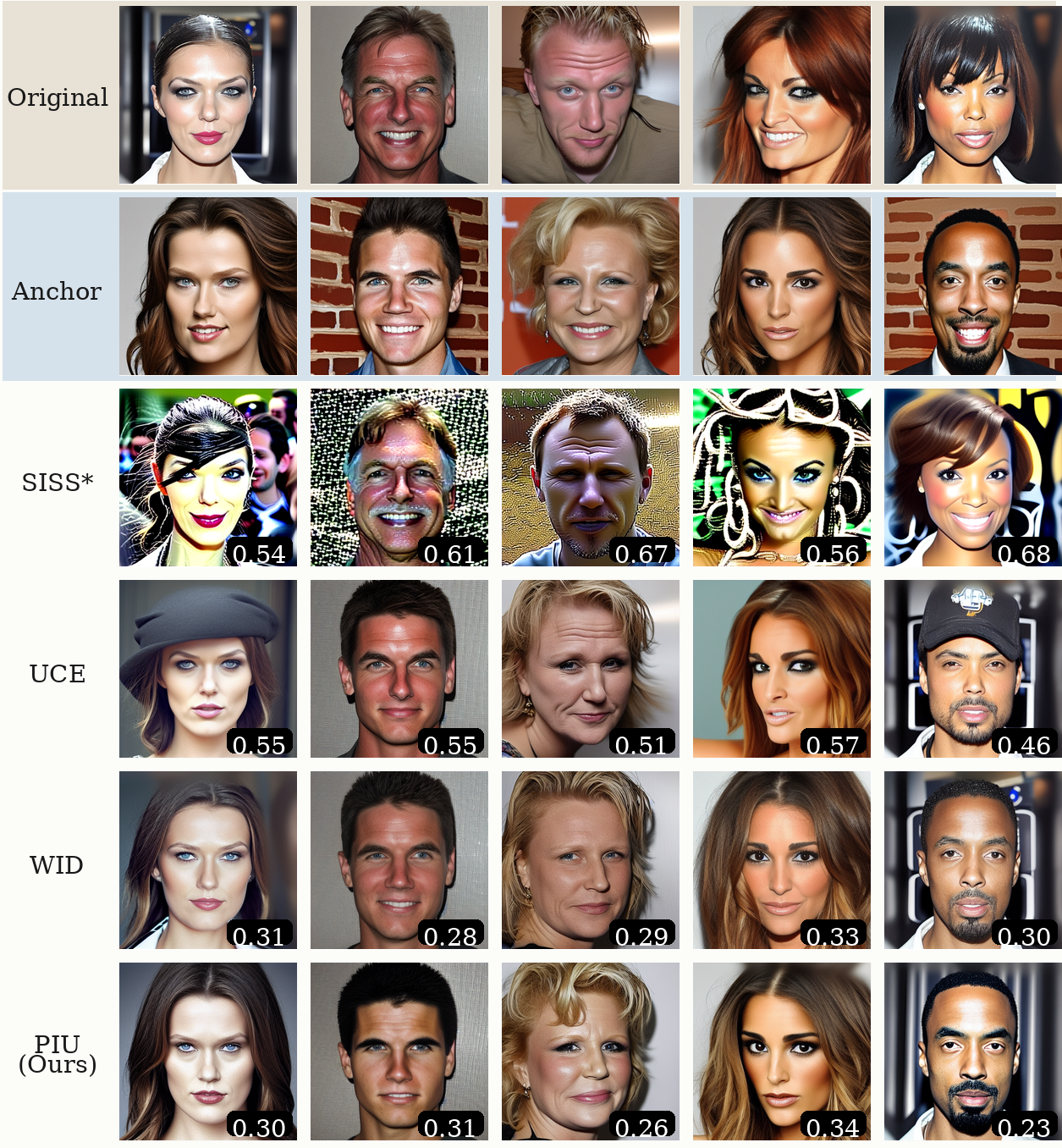}
    \caption{\textbf{Images generated by Arc2Face before (first row) and after unlearning with different methods (third row onward).} All samples use fixed original ID embeddings and initial noise. Unlearning shifts the output from the original identity toward the facial characteristics of the synthetic anchor identity (second row), apart from the anchorless SISS method. For each unlearned image, the cosine similarity to the original identity is also reported.}
    \label{fig:forget_results}
    \vspace{-4mm}
\end{figure}

PIU also demonstrate strong performance in preserving face recognition quality for both forget- and retain-conditioned generated samples, as measured by eDIFFIQA. The absolute change in kernel distance ($\Delta \text{KD}$) shows that the retain set experiences negligible drift after PIU unlearning (lowest among all methods), due to the preservation-enforcing term in our loss. In contrast, $\Delta \text{KD}$ for the forget set is higher for PIU than for other methods, driven by the negative guidance scale $\eta$ (see Tables~\ref{tab:ablation_simplified} and \ref{tab:neg_scale_sweep}), and the anchor proximity value $\tau$ (see Table~\ref{tab:anchor_sweep}).

\begin{figure}[t]
        \centering
        \includegraphics[width=0.85\linewidth]{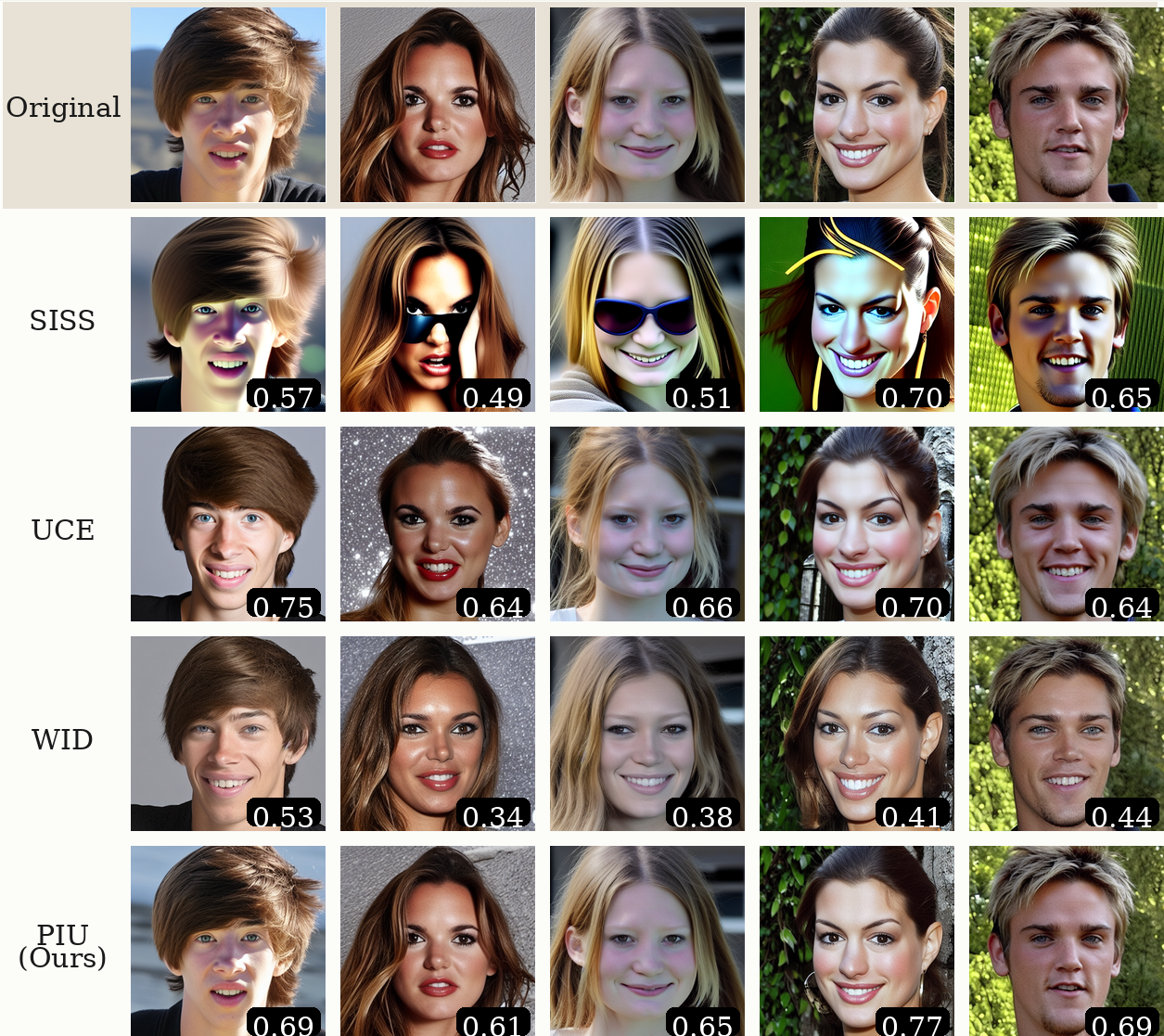}
        \caption{\textbf{Arc2Face samples of identities held-out from unlearning.} The same conditioning embeddings and noise are used for both the original and unlearned models, enabling a comparison of quality and identity preservation on non-target identities. Reported is the cosine similarity to the original non-target identity.}
        \label{fig:retain_results}
        \vspace{-4mm}
\end{figure}

\vspace{-2mm}
\paragraph{Qualitative results.}
%\label{subsec:qualitative_results}
The qualitative assessment of PIU in Figure~\ref{fig:forget_results} shows that post-unlearning samples effectively shift the original visual features toward those of their selected anchors. Both PIU and WID successfully achieve visual identity replacement. In contrast, UCE retains some of the original features (e.g. skin color), while the anchorless SISS significantly degrades visual quality and realism. Importantly, non-unlearned retain samples for a given forget–anchor pair in Figure~\ref{fig:retain_results} show that post-PIU outputs remain nearly identical to those generated prior to unlearning. Differently, WID exhibits clear anchor contamination in retain samples, as it lacks any preservation constraint. UCE introduces minor variations, while SISS does not alter identity but noticeably degrades visual realism.

\subsection{Ablation Studies}
\label{subsec:ablations}
For efficiency, ablation experiments are conducted on a fixed subset of $10$ forget identities, reporting the metrics most relevant to each component under study.
Table~\ref{tab:ablation_simplified} summarizes the performance as each component of PIU is introduced incrementally to isolate the contribution of each design choice. The naive unlearning baseline  (i.e., $\lambda=0$, $\eta=0$) shows that optimizing only the forget objective leads to poor identity preservation. Incorporating the preservation term $\mathcal{L}_{\mathrm{retain}}$ with weight $\lambda=10$ stabilizes the retain set, maintaining a higher average ISM. Furthermore, the negative guidance scale $\eta$ strengthens the unlearning effect, yielding a clear reduction in forget ISM and a substantial improvement in SRK. The ablation demonstrates that both components, $\lambda$ and $\eta$, are necessary to achieve effective unlearning while preserving model quality. The KD metric exhibits only minor changes from pre- to post-unlearning, which are primarily confined to the targeted forget identities, while the retain set remains largely unaffected.

\begin{table}[t]
\centering
% \caption{\textbf{Ablation of PIU components.} We compare a naive unlearning baseline, i.e. fine-tuning with only $\hat{\epsilon}_{\mathrm{forget}}$ using $\lambda{=}0$ and $\eta{=}0$, with variants that add the preservation term and negative guidance, and train only surgical layers. %, which reduces to $\hat{\epsilon}_{\mathrm{forget}} = \epsilon_{\theta^*}(z_t, t, c_a)$.}
% }
\caption{\textbf{Ablation of PIU components.} We incrementally introduce the preservation term, negative guidance, and surgical fine-tuning to a naive unlearning baseline ($\lambda{=}0$ and $\eta{=}0$). As shown, all components are essential to achieve effective identity removal without degrading quality and non-target identity consistency.}
\label{tab:ablation_simplified}
%\small
%\renewcommand{\arraystretch}{1.3}
%\setlength{\tabcolsep}{5pt}
\resizebox{\columnwidth}{!}{%
\begin{tabular}{l|cc|cc|c}
\hline

& \multicolumn{2}{c|}{\textbf{ISM}}
& \multicolumn{2}{c|}{\textbf{$\Delta$KD}}
& \textbf{SRK} \\
\cline{2-3} \cline{4-5} \cline{6-6}
\textbf{Variant} & Forget $\downarrow$
& Retain $\uparrow$
& Forget $\downarrow$
& Retain $\downarrow$
& $\uparrow$ \\
\hline
Baseline 
& 0.760 & {0.734} & -- & -- & 0.99 \\
\hline
Naive %($\lambda=0$, $\eta=0$) 
& \textbf{0.250} & 0.410 & \underline{7.713} & 4.331 & 55.6 \\
+ Preservation %$(\lambda=10)$
& 0.564 & \textbf{0.716} & \textbf{5.873} & \underline{0.284} & 1.6 \\
+ Neg. guidance %$(\eta=1.5)$
& {0.298} & \underline{0.714} & 8.542 & \textbf{0.230} & \underline{74.1} \\
+ Surgical training
& \underline{0.280} & 0.709 & 9.431 & \textbf{0.230} & \textbf{90.1} \\
\hline
\end{tabular}%
}
%\vspace{-4mm}
\end{table}

\vspace{-2mm}
\paragraph{Effect of the preservation weight \boldmath{$\lambda$}.}
The preservation weight $\lambda$ controls the trade-off between effective unlearning and preservation of the pretrained identity prior (Table~\ref{tab:pres_weight_sweep}). Without preservation ($\lambda = 0$), retain ISM drops markedly, as the model tends to collapse and redirect non-target identities toward the anchor, which is also reflected in higher Anchor ISM. Increasing $\lambda$ stabilizes retain generations and prevents this collapse at the cost of weaker forgetting, with higher forget ISM and lower SRK. We find that intermediate values like $\lambda = 10$ provide the best balance between identity removal and preservation of generative capabilities.

\begin{table}[t]
\centering
%\caption{\textbf{Effect of the preservation weight \boldmath{$\lambda$}.} Intermediate values provide the best balance of forgetting and retain preservation.}
\caption{\textbf{Effect of the preservation weight \boldmath{$\lambda$}.} Increasing $\lambda$ stabilizes model degradation but weakens unlearning. Intermediate values yield the best balance of forgetting and retain preservation.}
\label{tab:pres_weight_sweep}
%\footnotesize
%\renewcommand{\arraystretch}{1.1}
%\setlength{\tabcolsep}{4pt}
\resizebox{\columnwidth}{!}{%
\begin{tabular}{l|cc|c|c|cc}
\hline

& \multicolumn{2}{c|}{\textbf{ISM}}
& \textbf{Anchor ISM}
& \textbf{SRK}
& \multicolumn{2}{c}{\textbf{$\Delta$KD}} \\
\cline{2-3} \cline{4-4} \cline{5-5} \cline{6-7}
\boldmath{$\lambda$} & Forget $\downarrow$
& Retain $\uparrow$
& Retain $\downarrow$
& $\uparrow$
& Forget $\downarrow$
& Retain $\downarrow$ \\
\hline
0  & \textbf{0.048} & 0.229 & 0.395 & 11.2 & 11.601 & 13.720 \\
5  & \underline{0.213} & 0.700 & 0.038 & \textbf{90.3} & \textbf{8.596} & 0.321 \\	
10 & 0.280 & \underline{0.709} & \underline{0.022} & \underline{90.1} & 9.431 & \underline{0.230} \\
20 & 0.376 & \textbf{0.712} & \textbf{0.016} & 27.2 & \underline{9.266} & \textbf{0.193} \\
\hline
\end{tabular}
}
\vspace{-4mm}
\end{table}

\vspace{-4mm}
\paragraph{Effect of anchor selection.}
We analyze the impact of anchor proximity by varying the target cosine similarity threshold $\tau$ used to select the anchor identity. For this ablation, we use a batch size of  $16$,  
and evaluate on $20$ forget identities, to assess this choice over a broader identity set. Results in Table~\ref{tab:anchor_sweep} and samples in the supplementary material (Fig.~\ref{fig:prox_sensitivity}) show a clear trade-off: as the anchor moves farther from the forget identity, forgetting becomes stronger, as reflected by lower forget ISM, but preservation degrades, with lower retain ISM and lower SRK. In this sense, $\tau=0.1$ is the strongest setting from a pure unlearning perspective, but it also leads to worse image quality. By contrast, $\tau=0.2$ achieves the best retain $\Delta$KD and maintains a more balanced forget $\Delta$KD, yielding the most favorable overall trade-off between effective identity removal, preservation, and generation quality. For $\tau = 0.3$, the anchor becomes increasingly close to the forget identity, which weakens forgetting and reduces SRK. Larger thresholds like $\tau=0.4$ and $\tau=0.5$ are less reliable, since too similar or even same-identity samples can be selected as anchors, as shown in the supplementary material.

\begin{table}[t]
\centering
% \caption{\textbf{Effect of anchor proximity}. $\tau=0.2$ provides the best trade-off between forgetting, preservation, and image quality.}
\caption{\textbf{Effect of anchor proximity \boldmath{$\tau$}.} Lower values maximize unlearning at the cost of image quality, while larger values weaken identity removal. $\tau=0.2$ provides a balance between identity forgetting, non-target preservation, and image quality.}
\label{tab:anchor_sweep}
%\small
\resizebox{\columnwidth}{!}{
\begin{tabular}{l|cc|cc|c}
\hline

& \multicolumn{2}{c|}{\textbf{ISM}}
& \multicolumn{2}{c|}{\textbf{$\Delta$KD}}
& \textbf{SRK} \\
\cline{2-3} \cline{4-5}
\textbf{Anchor} & Forget $\downarrow$
& Retain $\uparrow$
& Forget $\downarrow$
& Retain $\downarrow$
&  $\uparrow$ \\ 
\hline
$\tau$ = 0.1 & \textbf{0.289} & 0.712 & {7.837} & 0.468 & \textbf{65.82} \\
$\tau$ = 0.2 & \underline{0.392} & \underline{0.713} & 7.698 & \textbf{0.269} & \underline{46.57} \\
$\tau$ = 0.3 & 0.463 & \textbf{0.716} & \textbf{5.372} & 0.464 & 23.92 \\
$\tau$ = 0.4 & 0.449 & 0.714 & \underline{6.450} & \underline{0.391} & 42.15 \\
$\tau$ = 0.5 & 0.451 & \underline{0.713} & 6.737 & 0.482 & 42.68 \\
\hline
\end{tabular}
}
\vspace{-2mm}
\end{table}

\begin{figure}[t]
    \centering
    \includegraphics[width=0.85\linewidth]{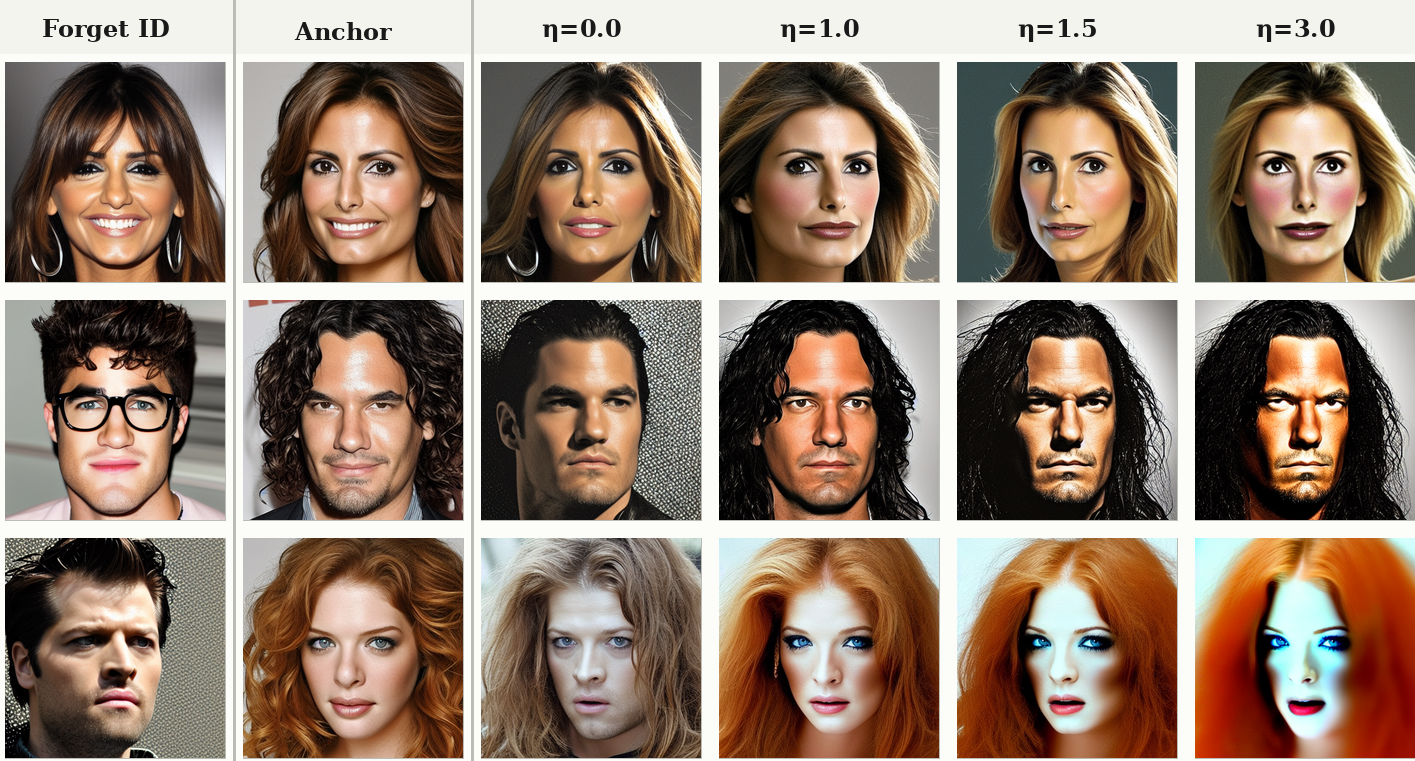}
\caption{\textbf{Effect of the negative guidance scale \boldmath{$\eta$}.} Increasing $\eta$ strengthens the shift away from original identities, but higher values degrade visual quality. With $\eta=0$, samples preserve the highest quality, but retain recognizable traits of original identities.}    \label{fig:neg_scale_sensitivity}
%\vspace{-2mm}
\end{figure}

\vspace{-4mm}
\paragraph{Effect of the negative guidance scale \boldmath{$\eta$}.}
The negative guidance term pushes the predicted noise away from the embeddings of the identities to be forgotten (see Eq.~\ref{eq:epsilon_forget}). Subtracting the anchor-conditioned prediction from the forget-conditioned one penalizes components that remain aligned with the original identity.
When $\eta=0$, the procedure reduces to matching the anchor prediction, encouraging generations that resemble the anchor. However, as seen in Table~\ref{tab:neg_scale_sweep}) and Figure~\ref{fig:neg_scale_sensitivity}, this does not actively suppress the underlying original identity. ISM with respect to the original identity remains high, leading to unchanged similarity-based matching and poor SRK performance. 
Increasing $\eta$ strengthens the unlearning effect, resulting in lower forget ISM and improved SRK, but can degrade generation quality with larger values, as reflected by increased KD scores.
We find that $\eta=1$ and $\eta=1.5$ provide the best trade-offs. Although $\eta=1.5$ gives stronger unlearning metrics, we observe that  $\eta=1$ with longer training reaches comparable unlearning while preserving quality more reliably. Indeed, on the same $10$ identities with $400$ training steps, $\eta=1$ reaches SRK $=90.1$ and forget ISM $=0.306$, while $\eta=1.5$ reaches SRK $=90.7$ and forget ISM $=0.247$, with similar behavior in the remaining metrics. Thus, we use $\eta=1$ as our default for results in Table~\ref{tab:unlearning_results}.

% \begin{table}[t]
% \caption{\textbf{Effect of the negative guidance scale \boldmath{$\eta$}.} Setting $\eta=0$ is insufficient to remove the target identity. Increasing $\eta$ improves forgetting and SRK performance, but large values may degrade the  $\Delta$KD and eDIFFIQA quality of forget-conditioned generations.}
% \label{tab:neg_scale_sweep}
% \centering
% % \small
% \setlength{\tabcolsep}{2pt}
% \resizebox{\linewidth}{!}{
% \begin{tabular}{l|cc|cc|cc|cc}
% \hline
% & \multicolumn{2}{c|}{\textbf{ISM}}
% & \multicolumn{2}{c|}{\textbf{SRK Metrics}}
% & \multicolumn{2}{c|}{\textbf{$\Delta$KD}}
% & \multicolumn{2}{c}{\textbf{eDIFFIQA}} \\
% \cline{2-3} \cline{4-5} \cline{6-7} \cline{8-9}
% \boldmath{$\eta$}
% & Forget $\downarrow$
% & Retain $\uparrow$
% & AccU $\downarrow$
% & SRK $\uparrow$
% & Forget $\downarrow$
% & Retain $\downarrow$
% & Forget $\uparrow$
% & Retain $\uparrow$ \\
% \hline 
% 0.0 
% & 0.564 & 0.716
% & 0.848 & 1.6 
% & \textbf{5.873} & \textbf{0.284}
% & \underline{0.745} & 0.747 \\

% 1.0 
% & 0.341 & \textbf{0.721}
% & 0.104 & 82.1 
% & \underline{9.047} & \underline{0.377}
% & \textbf{0.746} & \textbf{0.759} \\

% 1.5 
% & \underline{0.283} & \underline{0.720}
% & \underline{0.088} & \underline{90.1}
% & 9.578 & 0.431 
% & 0.732 & \underline{0.757}  \\

% 3.0 
% & \textbf{0.185} & 0.694 
% & \textbf{0.024} & \textbf{90.4}
% & 13.502 & 0.383 
% & 0.672 & 0.751  \\
% \hline
% \end{tabular}
% }
% \vspace{-4mm}
% \end{table}

\begin{table}[t]
\caption{\textbf{Effect of the negative guidance scale \boldmath{$\eta$}.} Setting $\eta=0$ is insufficient to remove the target identity. Increasing $\eta$ improves forgetting and SRK performance, but large values may degrade the $\Delta$KD and eDIFFIQA quality of forget-conditioned generations.}
\label{tab:neg_scale_sweep}
\centering
% \small
\setlength{\tabcolsep}{3pt}
\resizebox{\linewidth}{!}{
\begin{tabular}{l|cc|c|cc|cc}
\hline
& \multicolumn{2}{c|}{\textbf{ISM}}
& \textbf{SRK}
& \multicolumn{2}{c|}{\textbf{$\Delta$KD}}
& \multicolumn{2}{c}{\textbf{eDIFFIQA}} \\
\cline{2-3} \cline{5-6} \cline{7-8} 
\boldmath{$\eta$}
& Forget $\downarrow$
& Retain $\uparrow$
& $\uparrow$
& Forget $\downarrow$
& Retain $\downarrow$
& Forget $\uparrow$
& Retain $\uparrow$ \\
\hline 
0.0 
& 0.564 & 0.716
& 1.6 
& \textbf{5.873} & \textbf{0.284}
& \underline{0.745} & 0.747 \\

1.0 
& 0.341 & \textbf{0.721}
& 82.1 
& \underline{9.047} & \underline{0.377}
& \textbf{0.746} & \textbf{0.759} \\

1.5 
& \underline{0.283} & \underline{0.720}
& \underline{90.1}
& 9.578 & 0.431 
& 0.732 & \underline{0.757}  \\

3.0 
& \textbf{0.185} & 0.694 
& \textbf{90.4}
& 13.502 & 0.383 
& 0.672 & 0.751  \\
\hline
\end{tabular}
}
\vspace{-4mm}
\end{table}

\vspace{-4mm}
\paragraph{Effect of fine-tuned layers.}
Table~\ref{tab:surgical_balanced} compares three fine-tuning regimes: the full network, all cross-attention (CA) layers, and a surgical subset of CA layers. Fine-tuning the full model yields the weakest results, with unstable training and degraded quality for both forget and retain identities, reflected in consistent drops across all reported metrics. In contrast, both CA-based strategies achieve strong unlearning, with low forget ISM and high SRK, while updating only \qty{5.11}{\percent} and \qty{4.29}{\percent} of the parameters for full CA and surgical CA, respectively. Their difference is more subtle: full CA fine-tuning shows more stable KD behavior on forget samples, whereas surgical fine-tuning achieves better SRK while maintaining comparable ISM and image quality.

\begin{table}[t]
\centering
\caption{\textbf{Effect of fine-tuned layers.} Restricting PIU to Cross-Attention (CA) layers, especially the surgical subset, yields the best trade-off between unlearning, preservation, and efficiency.}
\label{tab:surgical_balanced}
%\small
\setlength{\tabcolsep}{2pt}
\resizebox{\columnwidth}{!}{%
\begin{tabular}{l|cc|cc|c|cc}
\hline
 %& Params 
\textbf{Tuning} & \multicolumn{2}{c|}{\textbf{ISM}}
& \multicolumn{2}{c|}{\textbf{$\Delta$KD}}
& \textbf{SRK} 
& \multicolumn{2}{c}{\textbf{eDIFFIQA}} \\
% \cline{2-3} \cline{4-5}  \cline{7-8}
%&
\textbf{(Parameters)}  & Forget $\downarrow$ & Retain $\uparrow$
& Forget $\downarrow$ & Retain $\downarrow$
& $\uparrow$
& Forget $\uparrow$ & Retain $\uparrow$ \\
\hline
Full (900M)  
%& 900M (100\%)  
& \textbf{0.134} & 0.294 
& 16.950 & 20.293
& 51.9 
& 0.653 & 0.543 \\

Full CA (44M)
%& 44M (5.11\%)  
& 0.298 & \textbf{0.714} 
& \textbf{8.542} & \textbf{0.230}
& \underline{74.1} 
& \underline{0.726} & \underline{0.752} \\

Surgical (\textbf{37M})
%& 37M (4.29\%)  
& \underline{0.280} & \underline{0.709}
& \underline{9.431} & \textbf{0.230}
& \textbf{90.1}
& \textbf{0.729} & \textbf{0.753} \\
\hline
\end{tabular}
}
\vspace{-2mm}
\end{table}

\section{Conclusion}

In this paper, we introduced PIU, a proximity-guided framework for identity unlearning in ID-conditioned diffusion models. By combining anchor-guided identity replacement, preservation-aware optimization, and localized fine-tuning of identity-sensitive layers, PIU achieves effective unlearning of target identities while maintaining realism and identity consistency for retained identities. Extensive experiments on Arc2Face~\cite{papantoniou2024arc2face} demonstrate that PIU consistently delivers the best overall trade-off between forgetting strength, non-target preservation, and image quality, surpassing prior methods for unlearning in the ID-conditioned setting. 
{%\color{blue}
While our study focuses on single-identity unlearning in Arc2Face, the proposed PIU framework is general and opens directions for future work, including extensions to other ID-conditioned generative models and sequential unlearning scenarios, for which localized fine-tuning and proximity-based anchors minimize catastrophic forgetting.}

%While our study focuses on single-identity unlearning in Arc2Face, the proposed framework is general and opens promising directions for future work, including extensions to broader ID-conditioned generative architectures, multi-identity or sequential unlearning scenarios, and richer evaluation protocols for privacy and robustness. 

% Overall, these findings establish PIU as an effective and reproducible approach for controlled identity removal.

% While our study focuses on single-identity unlearning in Arc2Face~\cite{papantoniou2024arc2face}, the proposed framework is general and opens promising directions for future work, including extensions to broader ID-conditioned generative architectures, richer evaluation protocols for privacy, and sequential unlearning scenarios, where PIU's localized fine-tuning and proximity-based anchors inherently minimize the weight drift that triggers catastrophic forgetting.

% \TODO{Multi-identity unlearning remains a critical open challenge due to catastrophic forgetting. However, for sequential unlearning, PIU possesses a distinct advantage. By restricting fine-tuning to surgical cross-attention layers (only 4.29\% of the network) and enforcing preservation loss, PIU inherently minimizes weight drift. Furthermore, our proximity-based anchor selection constrains the unlearning trajectory to localized manifold regions, preventing structural distortions that can trigger catastrophic forgetting. PIU also supports joint unlearning by processing multiple target-anchor pairs in the same batch.}

\section*{Acknowledgements}
Supported in parts by the Slovenian Research and Innovation Agency
(ARIS) through Research Programmes P2-0250 (B) “Metrology and Biometric Systems” and P2–0214 (A) “Computer Vision”, the ARIS Young Researcher Programme, and
the European Union’s Horizon Europe research and innovation programme
through the OnMoveID project under grant agreement No. 101225635.

{\small
\bibliographystyle{ieee}
\bibliography{bibliography}
}

\clearpage

% Supplementary first page
\twocolumn[
\begin{center}
    {\LARGE\bfseries PIU: Proximity-Guided Identity Unlearning in ID-Conditioned Diffusion Models\par}
    \vspace{0.4em}
    {\large Supplementary Material\par}
    \vspace{1.0em}
\end{center}
]

% \iffalse
\appendix
\section{Additional Experimentation Details}
\label{sec:supp_dataset}

\paragraph{Dataset Details.}
As explained in Section \ref{sec:experiments}, we use the CelebA-HQ~\cite{karras2018progressive}. Specifically we use the version available on the Hugging Face Hub\footnote{\url{https://huggingface.co/datasets/jxie/celeba-hq}}, with the snapshot at revision \texttt{7ecc6a45edfb5483ccf2f7df1035d298ffe7c76b}.
To perform unlearning, it is necessary to identify the identities present in the dataset. However, the referenced CelebA-HQ dataset provides only gender labels, with no identity annotations. For this reason we design a clustering strategy to automatically group samples belonging to the same identity.

We cluster the embedding space using DBSCAN, a density-based method that groups samples according to local similarity without requiring a predefined number of clusters. We use the Scikit-Learn implementation with cosine distance.

\begin{figure}[!ht]
    \centering
    \includegraphics[width=\linewidth]{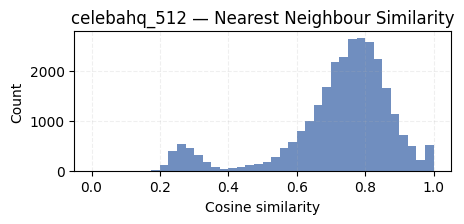}
    \caption{Distribution of the cosine similarity of each sample's nearest neighbor. Two evident modes appear for same- and similar-identity samples.}
    \label{fig:nearest_neighbor_sim}
\end{figure}

Figure~\ref{fig:nearest_neighbor_sim} shows the distribution of cosine similarity for all samples' nearest neighbors. Two distinct modes emerge, with peaks around $\tau \approx 0.8$ and $\tau \approx 0.25$. The high-similarity peak corresponds to samples of the same identity, while the lower peak captures ArcFace-similar but distinct individuals. Between these modes, a minimum appears around $\tau \approx 0.4$, providing a natural separation threshold.
Based on this observation, we set the cosine distance threshold to $\epsilon = 0.35$ and require a minimum of two samples per cluster. These choices define the DBSCAN configuration used for identity grouping.
This procedure produces $9{,}683$ clusters, which we use as identity labels during training. Manual inspection confirms that the resulting clusters are visually consistent. Empirically, we observe that similarities above $\tau > 0.6$ almost always correspond to the same individual, with only rare exceptions arising from varying lighting conditions or significant facial occlusions.

\begin{figure}[!ht]
    \centering
    \includegraphics[width=\linewidth]{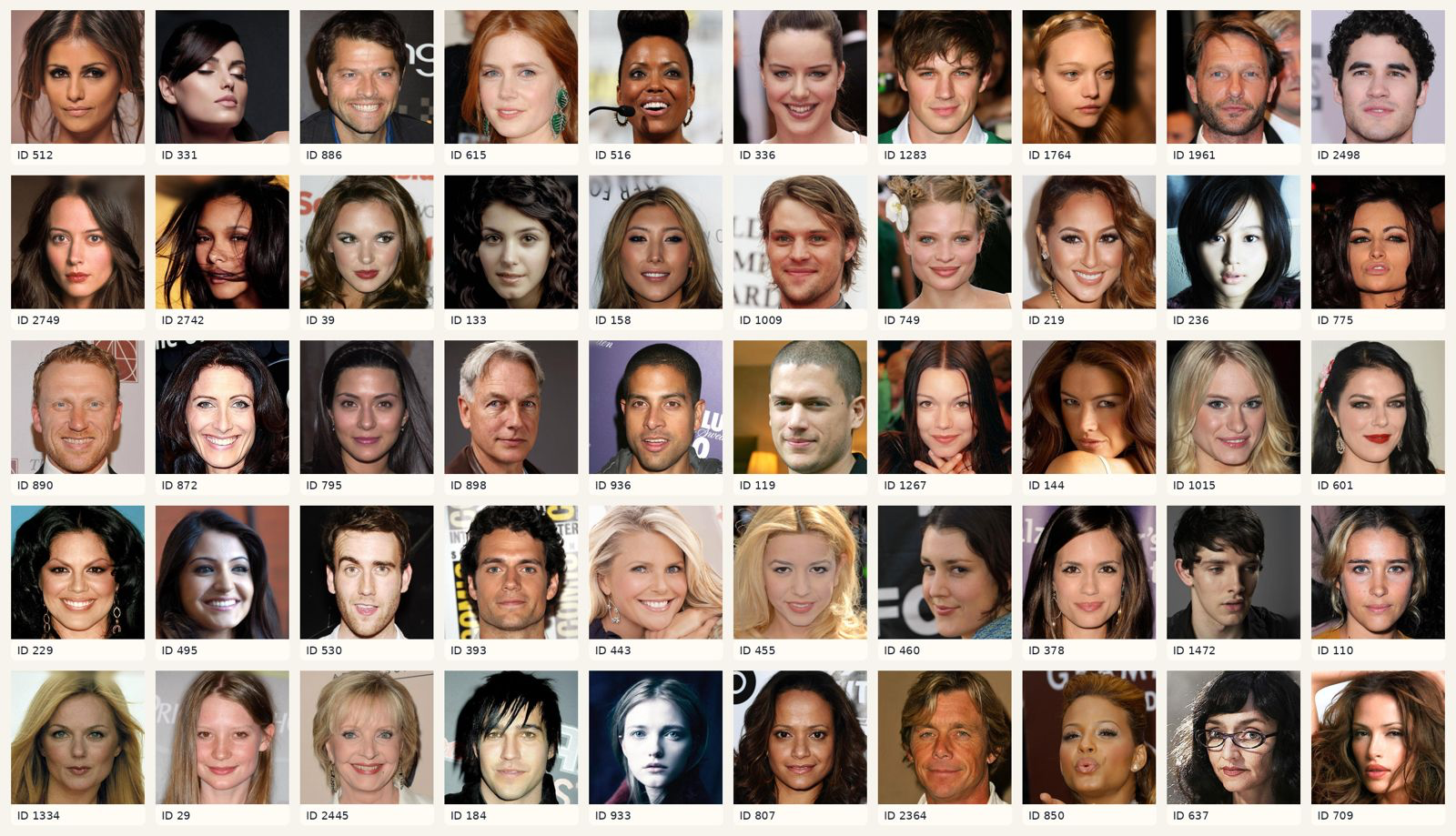}
    \caption{\textbf{Overview of the 50 manually curated target identities.} The selection enforces demographic diversity across gender and race alongside variations in facial pose, expression, and lighting.}
    \label{fig:used_identities}
    \vspace{-4mm}
\end{figure}

\paragraph{Manual Identity Selection for Unlearning.}
\label{sec:selected_unlearning_identities}
{%\color{blue}
As discussed in the main text, evaluating unlearning methods on a purely random sample from CelebA-HQ can result in an unbalanced evaluation set heavily skewed toward specific demographics. To ensure a rigorous and fair evaluation, we manually curated the $50$ target identities used in our benchmark comparisons. This curation process prioritized diverse representation across different demographic groups, specifically balancing race and gender, while retaining variance in facial poses, expressions, and lighting conditions to test the unlearning framework under realistic, in-the-wild variations. Figure~\ref{fig:used_identities} provides a visual overview of one representative image from each of the $50$ selected identities.}

% {\color{red}Our dataset is publicly available on Hugging Face at \url{https://huggingface.co/datasets/edgarcancinoe/celebahq_512_id_clusters} (for now this is private due to ICJB anonymity requirements).}

\begin{figure*}[!ht]
    \centering
    \includegraphics[width=\textwidth, trim={0 3 0 0}, clip]{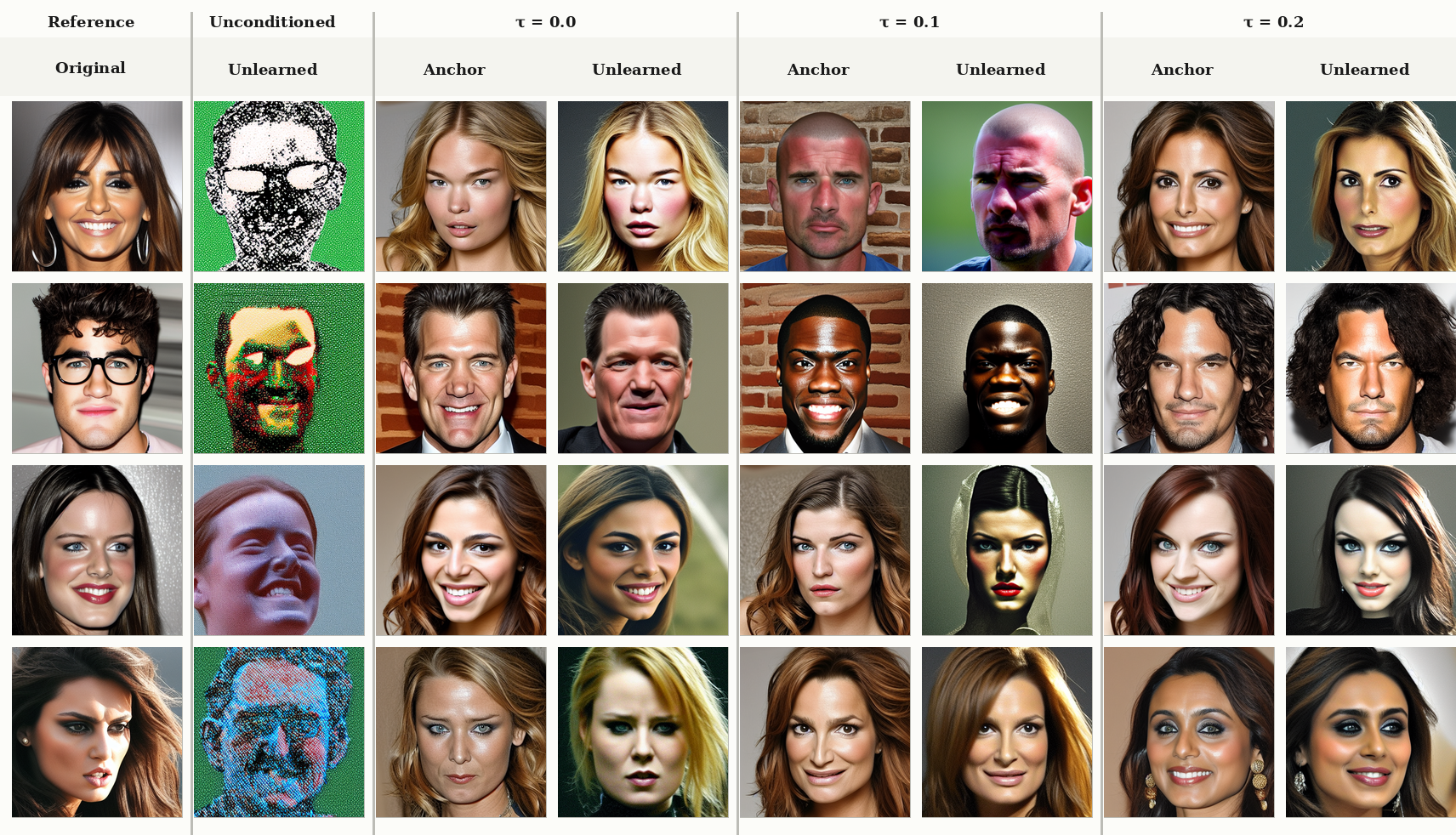}
    \vspace{-4mm}
\caption{\textbf{Qualitative comparison of PIU under different anchor selection strategies.} True unconditional guidance degrades realism and leads to unstable generations. In contrast, anchors with increasing proximity progressively improve visual quality and identity consistency.}    \label{fig:prox_sensitivity}
\end{figure*}
%\vspace{-4mm}

\paragraph{Details of Unlearning Metrics.}
\label{sec:supp_unlearning_metrics}

We assess identity removal using \emph{Identity Score Matching} (ISM)~\cite{van2023anti} and \emph{Selective Removal and Keep} (SRK)~\cite{shaheryar2025unlearn}. Although both metrics have appeared in prior work, their exact computation is often only briefly described or left implicit, especially in the context of ID-conditioned face generation. For reproducibility, we provide here the precise definitions and implementation details used in our evaluation. 

Let $\mu_k$ denote the ArcFace centroid of identity $k$, computed from its training images as in Equation~\ref{eq::centroid}. Given a set of generated samples $\{y_j^{(k)}\}_{j=1}^n$ associated with identity $k$, ISM is defined as the average cosine similarity between the ArcFace embedding of each generated image and its corresponding identity centroid:
\begin{equation}
\mathrm{ISM}(k)
=
\frac{1}{n}
\sum_{j=1}^n
\cos\big(\mathrm{ArcFace}(y_j^{(k)}), \mu_k\big).
\end{equation}
We use this metric both for forget and retain identities. For a forget identity, a low ISM indicates successful unlearning, since generated samples no longer remain close to its original identity representation in ArcFace space. For retain identities, on the contrary, a high ISM indicates that the identity prior is preserved after unlearning. We also use \emph{Selective Removal and Keep} (SRK)~\cite{shaheryar2025unlearn}, which evaluates forgetting and preservation jointly through nearest-centroid classification.

For a set of unlearned identities $\mathcal{U}$, we define the forget accuracy as
\begin{equation}
\mathrm{Acc}_U
=
\frac{1}{|\mathcal{U}|\,m}
\sum_{u \in \mathcal{U}}
\sum_{j=1}^m
\mathbf{1}\{\hat{y}^{(u)}_j = u\},
\end{equation}
where $\hat{y}^{(u)}_j$ is the identity predicted by nearest-centroid cosine similarity for the $j$-th generated sample associated with identity $u$. A successful unlearning method should make $\mathrm{Acc}_U$ low, ideally close to zero.

Similarly, for a set of retained identities $\mathcal{R}$, we define the retain accuracy as
\begin{equation}
\mathrm{Acc}_R
=
\frac{1}{|\mathcal{R}|\,m}
\sum_{r \in \mathcal{R}}
\sum_{j=1}^m
\mathbf{1}\{\hat{y}^{(r)}_j = r\}.
\end{equation}
A good preservation of the Arc2Face identity prior should keep $\mathrm{Acc}_R$ high.

These two quantities are combined into the SRK score:
\begin{equation}
\mathrm{SRK}
=
\frac{\mathrm{Acc}_R}{\mathrm{Acc}_U + \varepsilon},
\end{equation}
where $\varepsilon = 1e^{-2}$ is a small constant to avoid division by zero. Therefore, a higher SRK indicates better performance, as it reflects both effective removal of the target identity and preservation of the remaining ones. In this way, SRK complements ISM by checking not only whether an identity moves away from its original centroid, but also whether this shift causes confusion with other identities in the embedding space.

\section{Additional Results}
%\subsection{More Qualitative Results}
\label{sec:supp_qualtitative_results}

\begin{figure*}[!ht]
    \centering
    \includegraphics[width=\textwidth, trim={0 3 0 0}, clip]{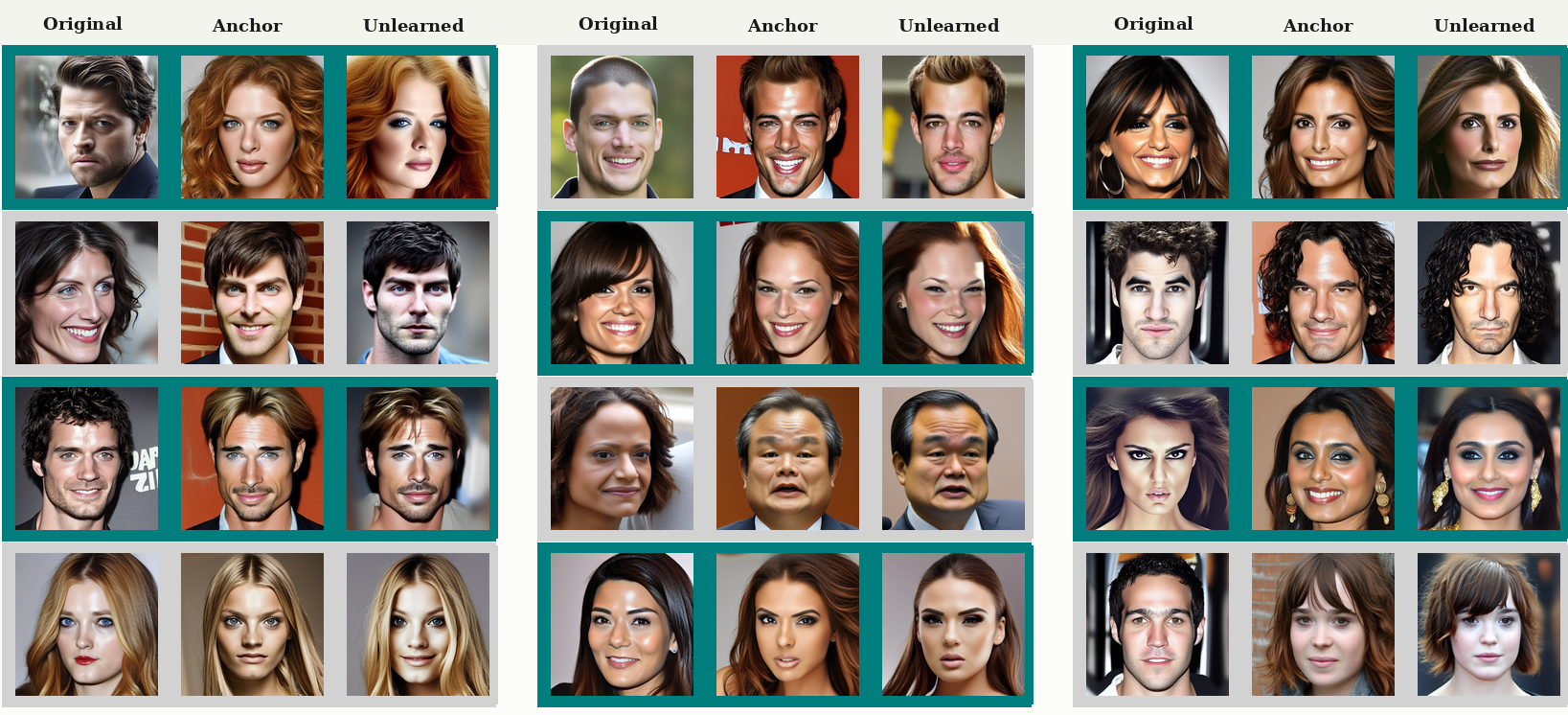}
    \vspace{-4mm}
\caption{Visualization of PIU results across 12 Original–Anchor–Unlearned triplets. Unlearned samples are generated using the same conditioning embeddings and initial noise as their corresponding original samples. The results show a consistent shift in identity, with outputs moving toward the facial characteristics of the selected anchors. The anchor images are also generated by Arc2Face.}    \label{fig:final_results}
\end{figure*}
% \vspace{-6mm}

\paragraph{Anchor Selection.}
Figure~\ref{fig:prox_sensitivity} illustrates the impact of anchor proximity on the unlearning behavior. Using true unconditional guidance results in degraded realism and unstable generations, confirming that the model relies on a meaningful identity prior. As the anchor proximity increases, the generated samples become progressively more stable and visually consistent, indicating a smoother transition in the identity space.
In practice, we select $\tau = 0.2$, as it provides the best trade-off between stable unlearning and high-quality post-unlearning generations when conditioning on the target identity.

\paragraph{Main results.}
The examples in figure \ref{fig:final_results} demonstrate that PIU achieves stable identity shifts across challenging settings, including cross-gender and cross-ethnicity transformations. By keeping the conditioning embeddings and initial noise fixed, the observed changes can be directly attributed to the unlearning procedure rather than stochastic variation.
Training is performed for 400 steps with $\eta=1.0$, using a batch size of 32 for both forget and retain samples per optimization step. Under these conditions, the method consistently drives the generated identities toward their assigned anchors while preserving the model's visual realism.

{%\color{blue}
\paragraph{Generalization to different identity spaces.}
\label{sec:generalization}

\begin{table}[!ht]
{%\color{blue}
\caption{\textbf{Comparison with other unlearning methods across different identity spaces.} In both ArcFace~\cite{deng2019arcface} and AdaFace~\cite{kim2022adaface} identity spaces, PIU achieves the strongest identity unlearning (lowest forget ISM and highest SRK). Bold denotes the best result for each metric; underline the second best.}
\label{tab:unlearning_results_adaface}
\centering
%\footnotesize
\setlength{\tabcolsep}{3pt}
\resizebox{\columnwidth}{!}{%
\begin{tabular}{l|cc|c|cc|c}
& \multicolumn{3}{c|}{\textbf{ArcFace}} & \multicolumn{3}{c}{\textbf{AdaFace}} \\
\hline 
 & \multicolumn{2}{c|}{\textbf{ISM}} & \textbf{SRK} $\uparrow$ & \multicolumn{2}{c|}{\textbf{ISM}} & \textbf{SRK} $\uparrow$ \\
\cline{2-3} \cline{4-4} \cline{5-6} \cline{7-7}
\textbf{Model} & Forget $\downarrow$ & Retain $\uparrow$ & & Forget $\downarrow$ & Retain $\downarrow$ &  \\
\hline
%Arc2Face \cite{papantoniou2024arc2face} & 0.78 & 0.75 & 0.99 & -- & -- & --  \\
% \hline
SISS \cite{alberti2025data}    & 0.59 & 0.58 & 0.98 & 0.59& 0.57& 0.98\\
UCE  \cite{gandikota2024unified}  & 0.59 & \underline{0.70} & 1.00 & 0.59& \underline{0.69} & 1.02\\
WID \cite{shaheryar2025unlearn}     & \underline{0.32} & 0.44 & \underline{49.28} & \underline{0.34} & 0.45& \underline{49.60}\\
\textbf{PIU} (Ours)& \textbf{0.31} &\textbf{0.72} & \textbf{88.96} & \textbf{0.31} & \textbf{0.71} & \textbf{92.64 }\\
\hline

\end{tabular}
}
%\vspace{-2mm}
} % color
\end{table}

Because Arc2Face utilizes ArcFace embeddings for identity conditioning, evaluating unlearning success within the same ArcFace latent space might introduce a potential circularity bias. To confirm that our method genuinely erases the identity features rather than just manipulating the ArcFace-specific manifold, we re-evaluated the identity related metrics (ISM and SRK) using a pretrained AdaFace recognition model~\cite{kim2022adaface}. As shown in Table~\ref{tab:unlearning_results_adaface}, the AdaFace-based results closely mirror our original findings with the ArcFace model~\cite{deng2019arcface}. This confirms that the identity unlearning achieved by PIU effectively transfers across different feature spaces and is not an artifact of the conditioning network.
}

% \section{Generalization to other generative models}
% \label{sec:generalization}
% \TODO{Performance on IDSync and their training dataset -- at least qualitative examples}

\section{Reproducing Unlearning Methods in Identity-Conditioned Diffusion}
\label{sec:supp_repr_methods}

In this section, we describe the reproduction and implementation details of the three baseline methods compared against our proposed method, PIU, in Section~\ref{subsec:quantitative_results}. We explain how each baseline was reconstructed, adapted to the Arc2Face framework, and evaluated. In particular, Subsection~\ref{subsec:siss_reproduction} covers our reproduction of \emph{Subtracted Importance Sampled Scores (SISS)}~\cite{alberti2025data} as the data-level unlearning baseline; Subsection~\ref{subsec:uce_reproduction} covers our reproduction of \emph{Unified Concept Editing in Diffusion Models (UCE)}~\cite{gandikota2024unified} as the concept-unlearning baseline; and Subsection~\ref{subsec:wid_reproduction} covers our reproduction of \emph{the Unlearn and Protect variant With ID-guidance (WID)}~\cite{shaheryar2025unlearn} as the identity-unlearning baseline.

\paragraph{Overview and Reproducibility Protocol.} 

All reproduced methods follow the same structure. We first briefly introduce the method and motivate its inclusion as a baseline for PIU. In the \emph{Original Method} subsection, we summarize the parts of the original formulation that are most relevant to our reproduction. In \emph{Our Adaptation}, we explain how the method is transferred to Arc2Face while remaining as faithful as possible to its original assumptions and design principles. The \emph{Experiments} subsection then describes the implementation details, evaluation protocol, and hyperparameter searches used to identify the strongest baseline configuration for fair comparison with PIU. Finally, in \emph{Results and Discussion}, we analyze the obtained results and compare them with the claims reported in the original papers.

\paragraph{Adaptation Challenges for Arc2Face.} 

A dedicated adaptation subsection is needed because none of the considered baselines was originally proposed for the setting studied in this work. Arc2Face is an identity-conditioned latent diffusion model, whose conditioning mechanism differ substantially from the unconditional or text-conditioned settings assumed by prior unlearning methods (Section ~\ref{subsec:preliminaries}). As a result, reproduction in our case requires more than direct implementation: the main components of each method must be reformulated so that they remain meaningful under identity conditioning. We therefore make explicit which parts can be transferred directly, which must be modified, and how these modifications were chosen to preserve the original intent of each method as closely as possible.

\subsection{Subtracted Importance Sampled Scores}
\label{subsec:siss_reproduction}

Among existing data unlearning methods for diffusion models, we choose to reproduce and adapt  Subtracted Importance Sampled Scores (SISS)~\cite{alberti2025data}. This choice is motivated by several factors. First, to the best of our knowledge, SISS is the most recent representative method in this line of work, and it empirically outperforms earlier data unlearning baselines such as EraseDiff~\cite{wu2025erasediff} on its reported benchmarks. Second, unlike earlier approaches, SISS is accompanied by a theoretical analysis of its objective, a stability analysis, and a well-documented experimental setup in the appendix. Third, the method is evaluated on unconditional diffusion models trained on CelebA-HQ, which is particularly relevant to our setting, since CelebA-HQ is also the dataset underlying our experiments. For these reasons, SISS provides a strong starting point for identity-level data unlearning, although its original formulation is not directly applicable to Arc2Face because our model is identity-conditioned rather than unconditional.

\subsubsection{SISS Methodology}
We follow the notation of Alberti et al.~\cite{alberti2025data}. Let $X=\{x_1,x_2,\dots,x_n\}$ denote the dataset used to pretrain a diffusion model $\epsilon_\theta$, and let $A=\{a_1,a_2,\dots,a_k\}\subset X$ be the subset of samples to be forgotten, with $|X|=n$ and $|A|=k$. The objective is to fine-tune the pretrained model so as to approximate the model that would have been obtained had the forget set $A$ not been present during training, while retaining the generative behavior induced by the remaining data $X\setminus A$. In the original paper, this is formalized through two related fine-tuning objectives. 

The first formulation, which we refer to as \emph{SISS-No-IS}, is obtained by rewriting the standard diffusion training loss restricted to the retained set $X\setminus A$. Starting from the DDPM forward process
\[
q(x_t \mid x_0)=\mathcal{N}(x_t;\gamma_t x_0,\sigma_t I),
\]
the objective over the retained distribution can be expressed as
\begin{equation}
\label{eq:siss_nois}
\begin{aligned}
L_{X\setminus A}(\theta)
&= \frac{n}{n-k}\,
\mathbb{E}_{\substack{x \sim p_X\\ x_t \sim q(\cdot \mid x)}}
\Bigl\|
\frac{x_t-\gamma_t x}{\sigma_t} - \epsilon_\theta(x_t,t)
\Bigr\|_2^2 \\
&\quad - \frac{k}{n-k}\,
\mathbb{E}_{\substack{a \sim p_A\\ a_t \sim q(\cdot \mid a)}}
\Bigl\|
\frac{a_t-\gamma_t a}{\sigma_t} - \epsilon_\theta(a_t,t)
\Bigr\|_2^2 .
\end{aligned}
\end{equation}
The notation $L_{X\setminus A}$ emphasizes that this objective corresponds to the diffusion training loss that would ideally be optimized using only the retained data distribution. Equation~\eqref{eq:siss_nois} makes this target explicit by decomposing it into a preservation term, which maintains performance on the full training distribution, and a subtractive forgetting term, which removes the contribution of the forget set $A$. However, in practice, evaluating~\eqref{eq:siss_nois} requires two separate forward passes through the denoising network, one for samples from $X$ and one for samples from $A$, which increases both the computational and memory cost.

To address this limitation, Alberti et al.~\cite{alberti2025data} introduce an importance-sampling reformulation, which they call \emph{SISS}. The key idea is to replace the two separate expectations by a single expectation over a proposal mixture distribution $q_\lambda(m_t \mid x,a) := (1-\lambda)\,q(m_t \mid x) + \lambda\,q(m_t \mid a),$
where $\lambda \in [0,1]$ controls the trade-off between sampling around retained examples and forget examples. The full expression of the resulting objective $\ell_\lambda(\theta)$ is given in the original paper~\cite{alberti2025data}.

A central theoretical result in the paper is that this importance-sampled objective is equivalent to the original deletion loss. More precisely, the authors show that
\[
\ell_\lambda(\theta) = L_{X\setminus A}(\theta), \qquad \forall \lambda \in [0,1],
\]
and further prove that the corresponding gradient estimators are also equal in expectation. Therefore, SISS preserves the same optimization target as the original two-forward formulation. At the same time, by estimating both contributions from a single sampled noisy point $m_t$, it requires only one forward-backward pass through the denoising network, making it computationally more efficient than the SISS-No-IS formulation.

\subsubsection{Adapting SISS to Arc2Face}

The empirical results reported in~\cite{alberti2025data} further motivate the use of SISS in our setting. In particular, on CelebA-HQ with an unconditional DDPM, it's shown that both SISS with \(\lambda=0.5\) and SISS-No-IS achieve a favorable trade-off between forgetting strength and quality preservation, with SISS-No-IS yielding slightly stronger unlearning among the quality-preserving variants. They also evaluate the method on Stable Diffusion, showing that it extends to conditional models when the conditioning variable is fixed and reliably identifies the target to be removed. In that setting as well, SISS-No-IS remains slightly stronger empirically than the one-pass SISS estimator.

Although Arc2Face is not an unconditional diffusion model, its conditioning structure is highly constrained: the textual prompt is fixed, and generation is controlled solely through the identity embedding (see Section~\ref{subsec:preliminaries}). This places Arc2Face in the same regime considered in SISS for conditional models with fixed conditioning. In our case, the forget set, defined in Section~\ref{subsec:surgical_layers}, consists of the identity embeddings associated with the target identity together with their linear combinations, all of which consistently generate the same subject. We therefore treat this identity-conditioned set as \(A\) in the SISS formulation and adapt the SISS-No-IS objective accordingly.

We chose SISS-No-IS rather than the importance-sampled SISS objective for two reasons. First, our training setup is not constrained by computation, so the additional cost of two forward passes is acceptable. Second, SISS-No-IS directly optimizes the original two-term deletion objective, and the original paper reports it to be slightly stronger empirically in the identity-removal regime.

At the same time, the released implementation does not optimize the theoretical SISS-No-IS loss exactly in the scalar form of Eq.~\eqref{eq:siss_nois}. From our inspection of the official codebase, the retain and forget terms are evaluated separately, and the parameter update is constructed through a gradient-level subtraction in which the forgetting contribution is further rescaled using gradient norms. Thus, the practical implementation is better interpreted as an empirical optimization variant of the theoretical objective, rather than as a literal minimization of Eq.~\eqref{eq:siss_nois}.

To clarify this connection, we next rewrite Eq.~\eqref{eq:siss_nois} in a form that exposes more directly its preservation and forgetting components. This derivation, and the corresponding interpretation of the implementation, are not given explicitly in the original paper; they are part of our own analysis based on the published formulation and the released code. First of all, we rewrite the loss \ref{eq:siss_nois}:

\begin{equation}
\label{eq:siss_nois_weighted}
L_{\mathrm{SISS\text{-}No\text{-}IS}}(\theta)
=
w_x \,\mathcal{L}_x(\theta)
-
w_a \,\mathcal{L}_a(\theta),
\end{equation}
where
\begin{equation}
w_x := \frac{n}{n-k},
\qquad
w_a := \frac{k}{n-k},
\end{equation}
with
\begin{equation}
\mathcal{L}_x(\theta)
:=
\mathbb{E}_{\substack{x \sim p_X\\ x_t \sim q(\cdot \mid x)}}
\left[
\left\|
\frac{x_t-\gamma_t x}{\sigma_t}
-
\epsilon_\theta(x_t,t)
\right\|_2^2
\right],
\end{equation}
and
\begin{equation}
\mathcal{L}_a(\theta)
:=
\mathbb{E}_{\substack{a \sim p_A\\ a_t \sim q(\cdot \mid a)}}
\left[
\left\|
\frac{a_t-\gamma_t a}{\sigma_t}
-
\epsilon_\theta(a_t,t)
\right\|_2^2
\right].
\end{equation}
Equivalently, assuming $w_x \neq 0$,
\begin{equation}
L_{\mathrm{SISS\text{-}No\text{-}IS}}(\theta)
=
w_x
\left(
\mathcal{L}_x(\theta)
-
\frac{w_a}{w_x}\,\mathcal{L}_a(\theta)
\right),
\end{equation}
which yields the effective relative forgetting weight
\begin{equation}
\lambda_{\mathrm{eff}}
:=
\frac{w_a}{w_x}
=
\frac{k}{n}.
\end{equation}

At the gradient level, this gives
\begin{equation}
\nabla_\theta L_{\mathrm{SISS\text{-}No\text{-}IS}}(\theta)
=
w_x
\left(
\nabla_\theta \mathcal{L}_x(\theta)
-
\lambda_{\mathrm{eff}} \nabla_\theta \mathcal{L}_a(\theta)
\right).
\end{equation}
Since \(w_x>0\) is a global multiplicative constant, it does not change the optimization direction induced by the gradient, but only its overall magnitude. Therefore, from the perspective of first-order optimization, the practically relevant quantity is the relative weighting between the retain and forget terms, namely
\begin{equation}
\label{eq:practical_siss_direction}
\nabla_\theta \mathcal{L}_x(\theta)
-
\lambda_{\mathrm{eff}} \nabla_\theta \mathcal{L}_a(\theta).
\end{equation}

This is precisely the form suggested by the released implementation:
\begin{equation}
L_{\mathrm{SISS\text{-}No\text{-}IS}}(\theta)
\approx
\,\mathcal{L}_x(\theta)
-
\lambda_{\mathrm{eff}} \,\mathcal{L}_a(\theta),
\end{equation}
Rather than keeping the theoretical coefficient \(\lambda_{\mathrm{eff}}=k/n\) fixed, however, the code rescales the forgetting gradient through its norm. If
\(
g_x := \nabla_\theta \mathcal{L}_x
\)
and
\(
g_a := \nabla_\theta \mathcal{L}_a,
\)
the forgetting contribution is multiplied by
\begin{equation}
\lambda_{\mathrm{eff}}
:=
\frac{\alpha}{\|g_a\|_2},
\end{equation}
so that the final update direction becomes
\begin{equation}
\label{eq:empirical_siss_update}
g_x - \lambda_{\mathrm{eff}} g_a.
\end{equation}
In the original implementation, $\alpha$ is chosen so that the forget term has magnitude approximately \(10\%\) of the retain term, i.e.,
\begin{equation}
\alpha
\approx
0.1\,\|g_x\|_2,
\end{equation}
which implies
\begin{equation}
\lambda_{\mathrm{eff}}
\approx
0.1\,\frac{\|g_x\|_2}{\|g_a\|_2}.
\end{equation}
This can be interpreted as a norm-controlled approximation of Eq.~\eqref{eq:practical_siss_direction}, where the forgetting weight is determined not only by dataset proportions, but also by the instantaneous geometry of the two gradients.

Adapting this formulation to Arc2Face, we replace the unconditional DDPM objective by the identity-conditioned latent denoising loss. Let \(c_r\) denote a retain identity condition and \(c_f\) a forget identity condition, with noisy latent \(z_t\) defined as in Section~\ref{subsec:preliminaries}. We define
\begin{equation}
\label{eq:siss_arc2face_terms}
\begin{aligned}
\mathcal{L}_{\mathrm{retain}}(\theta)
&=
\mathbb{E}_{z_0,\epsilon,t,c_r}
\left[
\|\epsilon-\epsilon_\theta(z_t,t,c_r)\|_2^2
\right],\\
\mathcal{L}_{\mathrm{forget}}(\theta)
&=
\mathbb{E}_{z_0,\epsilon,t,c_f}
\left[
\|\epsilon-\epsilon_\theta(z_t,t,c_f)\|_2^2
\right].
\end{aligned}
\end{equation}
The corresponding SISS-style Arc2Face objective is then
\begin{equation}
\label{eq:siss_arc2face_objective}
\mathcal{L}_{\mathrm{Arc2Face\text{-}SISS}}(\theta)
=
\mathcal{L}_{\mathrm{retain}}(\theta)
-
\lambda_{\mathrm{eff}}\,
\mathcal{L}_{\mathrm{forget}}(\theta).
\end{equation}
In practice, following the released implementation, we optimize it through the gradient-level rule
\begin{equation}
\label{eq:siss_arc2face_grad}
\lambda_{\mathrm{eff}}
=
\frac{\alpha}
{\|\nabla_\theta \mathcal{L}_{\mathrm{forget}}\|_2}.
\end{equation}
We consider the adaptive variant $\alpha$ proportional to the retain gradient norm, therefore
\begin{equation}
\lambda_{\mathrm{eff}}
=
\beta
\frac{\|\nabla_\theta \mathcal{L}_{\mathrm{retain}}\|_2}
{\|\nabla_\theta \mathcal{L}_{\mathrm{forget}}\|_2},
\end{equation}
where \(\beta>0\) is a tunable hyperparameter.

\subsubsection{SISS Experiments}

The general implementation details, including dataset construction, training environment, hardware configuration, and evaluation metrics, follow Section~\ref{sec:experiments}. Here we only describe the choices specific to the SISS reproduction.

Unlike our proposed method, SISS operates directly on training samples. Therefore, we follow the original formulation and use latents encoded from real images, rather than sampling from Gaussian noise. This ensures consistency with the theoretical objective in Eq.~\eqref{eq:siss_nois}, where the expectations are defined over the data distribution.

Due to computational constraints, we conduct the reproduction on a reduced subset of identities. Instead of the full set of $10$ forget identities used in our ablation setting, we restrict the experiments to $5$ targets. Unless otherwise stated, we use a batch size of $64$ (implemented as $16$ with gradient accumulation over $4$ steps) and run $60$ update steps, following the short fine-tuning regime recommended in~\cite{alberti2025data}. Optimization is performed using AdamW with zero weight decay.

Following the recommendations of SISS~\cite{alberti2025data}, we take as our reference configuration a learning rate of $5\times 10^{-6}$ and a forgetting coefficient $\beta = 0.1$. However, we perform a small grid search to verify whether these hyperparameters remain appropriate in our identity-conditioned latent diffusion setting. In particular, the optimization in Arc2Face differs because generation takes place in latent space, so slightly larger learning rates may accelerate adaptation without necessarily destabilizing training. For this reason, we compare the reference value $5\times 10^{-6}$ with a more aggressive alternative, $1\times 10^{-5}$.

For the forgetting-strength coefficient, we evaluate
\[
\beta \in \{0.1,\;0.2,\;0.3,\;0.5\}.
\]
This range covers both the conservative regime recommended in the original work and progressively stronger forgetting regimes, allowing us to study the trade-off between identity removal and generation quality in our setting. All models are evaluated using the metrics introduced in Section~\ref{sec:experiments}, including the identity-removal metrics ISM and SRK, and the quality metrics $\Delta$KD and eDIFFIQA. We report results on both forget identities and retain identities, distinguishing between the seen retain training set and the unseen retain validation set.

\subsubsection{SISS Results and Discussion}

We first analyze the more aggressive learning rate setting, $1\times 10^{-5}$. As shown in Table~\ref{tab:siss_highlr_quality}, this value leads to a clear collapse in generation quality. The degradation is already evident in the realism metrics on retain identities, particularly in the KD-based distances for both seen and unseen retain samples. Qualitatively, generations become visibly corrupted and structurally unstable, especially for identities that should represent the retained distribution $X \setminus A$, indicating that the fine-tuning process no longer preserves the original generative behavior of the model.

This effect is further confirmed by the drop in eDIFFIQA on unseen retain identities. Since eDIFFIQA reflects whether generated faces remain sufficiently coherent and recognizable for face analysis, this deterioration indicates that the model is no longer producing reliable facial images even outside the forget set. For this reason, we do not include qualitative visualizations or a detailed $\beta$-wise analysis for the $1\times 10^{-5}$ setting, as the dominant conclusion is simply that this optimization regime is too unstable for our Arc2Face reproduction.

\begin{table}[ht]
\centering
\small
\resizebox{\columnwidth}{!}{
\begin{tabular}{l|c|c|c}
\hline
$\beta$ 
& $\Delta$ Seen KD $\downarrow$ 
& $\Delta$ Unseen KD $\downarrow$ 
& Unseen eDIFFIQA $\uparrow$ \\
\hline
$\beta=0.1$ & 10.210 & 10.372 & 0.526 \\
$\beta=0.2$ & 11.153 & 9.819 & 0.548 \\
$\beta=0.3$ & 10.256 & 9.516 & 0.592 \\
$\beta=0.5$ & 11.109 & 11.876 & 0.508 \\
\hline
\end{tabular}
}
\caption{Effect of using a higher learning rate ($1\times 10^{-5}$) in the SISS reproduction. Results are reported for different values of $\beta$. We report only retain-side realism metrics, since the main effect of this setting is a severe degradation of image quality and facial recognizability.}
\label{tab:siss_highlr_quality}
\end{table}

We now turn to the learning rate recommended by the original SISS implementation appendix \cite{alberti2025data}, namely $5\times 10^{-6}$. In contrast to the higher learning rate discussed above, this setting yields stable optimization and allows a more meaningful analysis of the effect of the forgetting coefficient $\beta$. The corresponding results are reported in Table~\ref{tab:siss_main_results}.

From the perspective of identity unlearning, the results show the expected monotonic trend: increasing $\beta$ generally strengthens the forgetting effect. This is most clearly reflected in the forget-side ISM, which decreases as $\beta$ grows, indicating that the generated samples become progressively less aligned with the target identity centroid. The same tendency is partially captured by SRK, which also improves for larger values of $\beta$, although the absolute SRK values remain low overall. In particular, when $\beta=0.1$, the reduction in forget ISM is still very limited, showing that the method performs only weak identity removal in this conservative regime. More noticeable forgetting emerges only for $\beta=0.3$ and $\beta=0.5$.

\begin{table*}[ht]
\centering
\small
\resizebox{\textwidth}{!}{
\begin{tabular}{l|c|c|c|c|c|c|c}
\hline
$\beta$
& Forget ISM $\downarrow$
& Retain Unseen ISM $\uparrow$
& SRK $\uparrow$
& $\Delta$ Forget KD $\downarrow$
& $\Delta$ Retain Unseen KD $\downarrow$
& Forget eDIFFIQA $\uparrow$
& Retain Unseen eDIFFIQA $\uparrow$ \\
\hline
Baseline & 0.775 & 0.726 & 0.990 & -- & -- & 0.763 & 0.770 \\
\hline
$\beta=0.1$ & 0.605 & 0.562 & 0.950 & 6.834 & 7.740 & 0.657 & 0.634 \\
$\beta=0.2$ & 0.576 & 0.568 & 0.964 & 7.401 & 7.067 & 0.660 & 0.626 \\
$\beta=0.3$ & 0.459 & 0.569 & 1.030 & 17.805  & 7.105 & 0.510 & 0.647 \\
$\beta=0.5$ & 0.326 & 0.572 & 1.621 & 18.244 & 7.706 & 0.434 & 0.650 \\
\hline
\end{tabular}
}
\caption{SISS reproduction on Arc2Face with learning rate $5\times 10^{-6}$ for different values of $\beta$. The baseline corresponds to the original Arc2Face model before unlearning. We report identity-removal metrics (ISM and SRK) together with realism and recognizability metrics (KD and eDIFFIQA).}
\label{tab:siss_main_results}
\end{table*}

\begin{figure}
        \centering
        \includegraphics[width=0.9\linewidth]{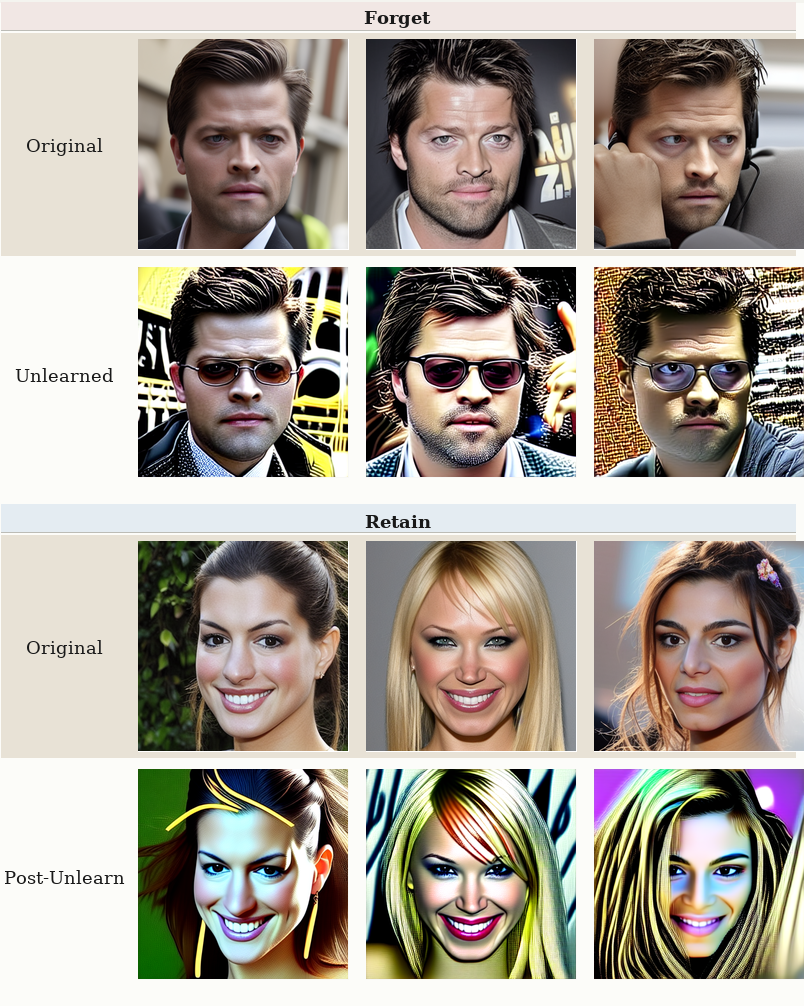}
        \caption{Qualitative results of the SISS reproduction on Arc2Face, with $\beta = 0.1$ and $lr = 5e^{-6}$. Row 1 shows baseline generations for the forget identity (ID 331), and Row 2 the corresponding generations after unlearning. Row 3 shows baseline generations for identities in the retain validation set, and Row 4 their generations after unlearning. Visual quality degrades substantially after fine-tuning: the effect is stronger for the forget identity, but remains severe for the retain identities as well.}
        \label{fig:siss_qualitative_results}
\end{figure}

However, this increase in forgetting strength comes at a substantial cost in generation quality. For larger values of $\beta$, both KD and eDIFFIQA deteriorate sharply, not only on the forget identities but also on the unseen retain identities. This indicates that the model does not merely become less faithful to the target identity; rather, its overall facial generation quality degrades significantly. A possible explanation is that, unlike our method, this SISS adaptation fine-tunes all model weights rather than restricting the update to the cross-attention layers. In our setting, such unrestricted fine-tuning appears to destabilize the generator more globally, which may explain the stronger degradation in image quality. This effect is especially problematic because it impacts the retain side as well: although the degradation is more severe for the forget set, it remains substantial for the retain set and is clearly worse than in our proposed method. At the same time, the eDIFFIQA values remain relatively high, suggesting that the generated faces are still broadly recognizable despite the loss in fidelity and realism. This is also consistent with the qualitative examples in Figure~\ref{fig:siss_qualitative_results}, where the generations remain identifiable but exhibit clear degradation, particularly in sharpness, structure, and overall realism. Even in the more favorable settings, the retain-side quality remains noticeably below the level achieved by our approach, both quantitatively and in the qualitative examples shown in Figure~\ref{fig:siss_qualitative_results}.

Taken together, these results suggest that SISS, at least in this adapted form, is not sufficiently effective for identity unlearning in identity-conditioned diffusion models such as Arc2Face. When $\beta$ is kept small, the method preserves generation quality reasonably well, but the amount of identity removal remains too limited to be competitive. When $\beta$ is increased to obtain stronger unlearning, the model rapidly loses image realism, leading to an unfavorable trade-off. Therefore, although the original SISS formulation is well motivated for data unlearning in unconditional or weakly conditioned settings, our experiments indicate that its direct extension to Arc2Face does not provide a satisfactory solution for targeted identity erasure.

Among the tested configurations, $\beta=0.1$ is the most reasonable operating point, as it is the only setting that does not severely compromise the generative quality of the model. Nevertheless, even this configuration yields only weak forgetting according to both ISM and SRK. We therefore conclude that the reproduced SISS baseline is ultimately too mild to achieve strong identity unlearning in our framework, while more aggressive hyperparameter choices destabilize the model and degrade the quality of the generated faces.

\subsection{Unified Concept Editing}
\label{subsec:uce_reproduction}

As discussed in Section~\ref{sec:RelatedWork}, concept unlearning has been one of the most explored directions for unlearning in diffusion models, especially in text-to-image settings. Within this literature, Unified Concept Editing (UCE)~\cite{gandikota2024unified} is a particularly relevant reference point for several reasons. First, it is one of the strongest and most established model-editing approaches for concept erasure, showing competitive or superior performance with respect to influential baselines such as ESD~\cite{gandikota2023erasing} and Ablating Concepts~\cite{kumari2023ablating}, that are similar to our method, while also preserving the quality of non-target generations. Second, UCE naturally supports simultaneous editing of multiple concepts, placing it in the same broader line of scalable multi-concept erasure methods. % as % MACE~\cite{lu2024mace}, 
% SepME~\cite{zhao2024separable}. 
Third, unlike gradient-based unlearning methods, UCE performs direct closed-form parameter editing without iterative fine-tuning, which makes it conceptually complementary to our own approach. This is especially relevant in our setting, since it allows us to test whether identity unlearning in Arc2Face can also benefit from a direct model-editing strategy rather than relying exclusively on fine-tuning. More broadly, UCE has become one of the standard references for adapting concept-removal methods to identity-related forgetting, as is also mentioned in GUIDE~\cite{seo2024generative} and Forget-Me~\cite{qi2025forget}.

\subsubsection{UCE Methodology}

As discussed in Section~\ref{subsec:preliminaries}, latent diffusion models incorporate conditioning information through cross-attention layers distributed throughout the U-Net. UCE~\cite{gandikota2024unified} operates directly on these layers, and more specifically on the linear projections that map conditioning embeddings into keys and values. Since the conditioning signal enters the denoising network through these projections, modifying them provides a direct mechanism to alter the association between a conditioning concept and its generated visual attributes.

Following the notation of~\cite{gandikota2024unified}, let \(c_i\) denote the embedding of a conditioning concept. At a given cross-attention layer, the corresponding key and value vectors are obtained as
\begin{equation}
k_i = W_k c_i, 
\qquad
v_i = W_v c_i,
\end{equation}
where \(W_k\) and \(W_v\) are the key and value projection matrices, respectively. These interact with the query representation \(q\), computed from the current latent features of the U-Net, to produce the cross-attention output
\begin{equation}
A \propto \mathrm{softmax}(q k_i^\top),
\qquad
O = A v_i.
\end{equation}
Thus, the conditioning influences the latent representation only through these key and value projections. UCE is based on the idea that editing these projections, both \(W_k\) and \(W_v\), is sufficient to suppress a target concept while preserving the behavior of the model on unrelated concepts.

To formalize this, the method defines a set \(E\) of concepts to edit, where each \(c_i \in E\) is a conditioning embedding corresponding to a concept to be erased. For each edited concept, a guide (or anchor) concept \(c_i^*\) is specified, representing the desired target behavior after editing. The corresponding target output is then defined through the original projection as
\begin{equation}
v_i^* = W_v^{\mathrm{old}} c_i^* .
\end{equation}
In the artist erasure setting considered in the original paper, \(c_i^*\) is not an identity-specific anchor, but rather a more generic target concept intended to remove the distinctive characteristics of the erased style. In addition, UCE introduces a set \(P\) of preservation concepts, whose outputs should remain unchanged after editing, in order to maintain generation quality on non-target concepts.

Given the edit set \(E\), the target outputs \(\{v_i^*\}\), and the preservation set \(P\), UCE updates the projection matrix \(W\) by solving the following least-squares objective:
\begin{equation}
\min_W
\sum_{c_i \in E} \|W c_i - v_i^*\|_2^2
+
\sum_{c_j \in P} \|W c_j - W^{\mathrm{old}} c_j\|_2^2.
\label{eq:uce_objective}
\end{equation}
As shown in~\cite{gandikota2024unified}, this objective admits the closed-form solution
\begin{equation}
\begin{split}
W = \left( \sum_{c_i \in E} v_i^* c_i^\top + \sum_{c_j \in P} W^{\mathrm{old}} c_j c_j^\top \right) \\
\times \left( \sum_{c_i \in E} c_i c_i^\top + \sum_{c_j \in P} c_j c_j^\top \right)^{-1}
\end{split}
\label{eq:uce_closed_form}
\end{equation}
This derivation is typically presented for the value projection, but the same closed-form update is applied to both \(W_v\) and \(W_k\), yielding a direct parameter edit of the cross-attention mechanism without iterative fine-tuning.

\subsubsection{Adapting UCE to Arc2Face}
For our reproducibility study and comparison in Section~\ref{subsec:quantitative_results}, we adapt UCE to the Arc2Face identity-conditioned setting, taking inspiration from the artist erasure experiments in the original paper. Although artist style removal is not identical to identity unlearning, both settings share the same core objective: suppressing a specific target concept while preserving performance on the remaining ones.

The main difference lies in the conditioning space, which in Arc2Face is identity-based rather than textual. Accordingly, the edited concepts are ArcFace identity embeddings mapped to the conditioning representations used by Arc2Face. Our forget set \(E\) is built from the same identity data used in our main method (Section~\ref{subsec:surgical_layers}), and includes multiple conditioning embeddings for the target identity together with their linear combinations. Thus, \(E\) represents a local region of the identity manifold associated with the target subject, and UCE is applied to redirect this region toward a fixed target (anchor).

The target construction is also modified. In the original UCE setting, each edited concept is redirected toward a generic guide concept. In our adaptation, all edited embeddings are redirected toward a single anchor identity. More precisely, for each \(c_i \in E\), we define the target concept \(c_i^*\) as the centroid of a selected anchor identity cluster, where the anchor is chosen following the procedure described in Section~\ref{subsec:threshold_optimized_anchor_selection}. Hence, unlike the original formulation, the target is not a generic semantic concept but a concrete identity-conditioned reference point.

We also adapt the construction of the preservation set \(P\). In Appendix D of~\cite{gandikota2024unified}, the artist erasure experiments show that a relatively large preservation set helps balance erasure strength and image quality. However, since Arc2Face is specialized to facial identity, using hundreds of raw preserve embeddings is less well motivated in our setting. Instead, we construct \(P\) from one normalized centroid per selected retain identity, yielding a compact and diverse set of preservation directions that covers a broader range of identities while avoiding over-representation of subjects with many highly similar embeddings in the retain split.

Once the edit set \(E\), the preservation set \(P\), and the anchor target have been specified, the implementation of our method follows closely the official UCE code released by the authors. In particular, we retain the same closed-form editing paradigm over the cross-attention key and value projections, while adapting the construction of the edited and preserved conditioning representations to the Arc2Face setting. Concretely, we use a weighted version of Eq.~\eqref{eq:uce_objective}, introducing coefficients \(\alpha_e > 0\) and \(\alpha_p > 0\) to control the relative strength of the unlearning and preservation terms, respectively, together with a regularization parameter \(\lambda > 0\) that stabilizes the update and keeps the edited weights close to the original ones. This yields the following objective:
\begin{equation}
\begin{aligned}
\min_W \quad & \alpha_e \sum_{c_i \in E} \|W c_i - v_i^*\|_2^2 \\
& + \alpha_p \sum_{c_j \in P} \|W c_j - W^{\mathrm{old}} c_j\|_2^2 \\
& + \lambda \|W - W^{\mathrm{old}}\|_F^2
\end{aligned}
\label{eq:uce_weighted_objective}
\end{equation}
The corresponding closed-form solution is
\begin{equation}
\begin{split}
W = \bigg( &\alpha_e \sum_{c_i \in E} v_i^* c_i^\top + \alpha_p \sum_{c_j \in P} W^{\mathrm{old}} c_j c_j^\top + \lambda W^{\mathrm{old}} \bigg) \\
\times \bigg( &\alpha_e \sum_{c_i \in E} c_i c_i^\top + \alpha_p \sum_{c_j \in P} c_j c_j^\top + \lambda I \bigg)^{-1}
\end{split}
\label{eq:uce_weighted_closed_form}
\end{equation}
which is the form implemented in our code for both \(W_v\) and \(W_k\).

\subsubsection{UCE Experiments}

The general implementation details, including dataset construction, experimental environment, and evaluation metrics, follow Section~\ref{sec:experiments}. UCE does not require iterative fine-tuning, as the update is given by the closed-form solution in Eq.~\eqref{eq:uce_weighted_closed_form}. As a result, the method is computationally lightweight in our setting: once the retain set \(P\) is precomputed, the unlearning edit for a single identity takes only a few seconds (at most $\sim$6 seconds in our experiments).

For consistency with the rest of our pipeline, we use the same forget and retain datasets as in Section~\ref{subsec:ablations}. However, due to the large number of hyperparameter configurations explored, we restrict the study to 5 forget identities instead of the full set of 10, enabling a broader search while maintaining a representative evaluation.

The main hyperparameters of UCE in our setting are \(\alpha_e\), \(\alpha_p\), and \(\lambda\) in Eq.~\eqref{eq:uce_weighted_objective}. Since the original paper and implementation do not provide clear recommended values, we first performed preliminary exploratory experiments over small ranges to identify stable regimes and define the final grid search.

These preliminary experiments revealed three consistent trends. First, effective forgetting required substantially larger values of \(\alpha_e\) than \(1\), since weaker edit weights were generally insufficient to produce a meaningful displacement toward the anchor identity. Second, \(\alpha_p\) had to remain comparatively smaller: values around \(1\) provided the most reasonable compromise between preservation and erasure, whereas larger preservation weights significantly weakened forgetting. Third, the regularization parameter \(\lambda\) was most effective in the range \((0,1)\), since stronger regularization made the update too conservative despite the relatively large values needed for \(\alpha_e\).

We also used these exploratory runs to compare two implementation choices: applying the closed-form edit only to the surgical cross-attention layers identified in Section~\ref{subsec:surgical_layers}, or to all cross-attention layers as in the original UCE formulation. We did not observe a substantial or systematic difference between the two variants. Therefore, in the final experiments we retained the all-layer version, which is both simpler and more faithful to the original method.

Based on these observations, we then performed a larger grid search over the most promising values:
\begin{gather*}
\alpha_e \in \{5, 10, 20, 50, 100\}, \\
\alpha_p \in \{0.1, 1, 5, 10\}, \\
\lambda \in \{0.3, 0.5, 0.7\}.
\end{gather*}
All configurations were evaluated using the same unlearning and quality metrics employed throughout the paper, allowing us to identify the strongest UCE baseline.

\subsubsection{UCE Results and Discussion}

We begin by analyzing the full grid search described above. A first important observation is that all 240 evaluated configurations yield exactly the same SRK score, \(0.9901\), with \(\mathrm{AccU}=1\) and \(\mathrm{AccR}=1\) in every case. This indicates that, under SRK, the method preserves identity recognition on the unseen retain validation set while failing to perform effective identity removal. We attribute this behavior to the fact that UCE does not explicitly enforce identity replacement, but rather pushes the target concept toward a more generic alternative. In the Arc2Face setting, this is not sufficient to break identity recognition in the sense captured by SRK.

Since SRK is uninformative in this regime, we focus on the trade-off between forget ISM and retain unseen ISM. Given the large number of configurations, results are summarized using boxplots, and our analysis is based on the median behavior of each hyperparameter.

\begin{figure}[ht]
    \centering
    \includegraphics[width=0.8\linewidth]{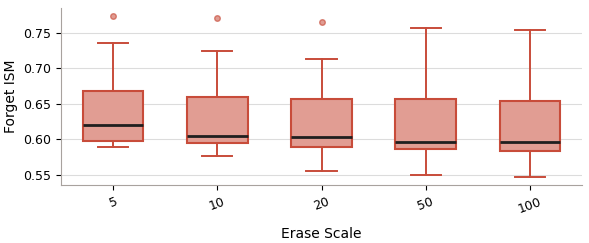}
    \caption{Forget ISM obtained across the grid search for different values of \(\alpha_e\). Each point corresponds to one experiment. Lower values indicate stronger identity removal.}
    \label{fig:uce_alphae_forget_ism}
\end{figure}

\begin{figure}[ht]
    \centering
    \includegraphics[width=0.8\linewidth]{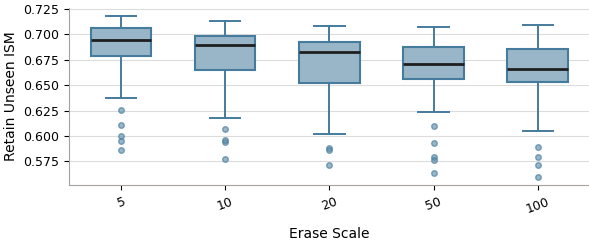}
    \caption{Retain unseen ISM obtained across the grid search for different values of \(\alpha_e\). Each point corresponds to one experiment. Higher values indicate better preservation of unseen retain identities.}
    \label{fig:uce_alphae_retain_ism}
\end{figure}

We first analyze the effect of the edit coefficient \(\alpha_e\), which directly controls the strength of unlearning. As shown in Figure~\ref{fig:uce_alphae_forget_ism}, the lowest median forget-ISM values are obtained for \(\alpha_e \in \{20, 50, 100\}\). Among these, \(\alpha_e=20\) achieves comparable median forgetting while exhibiting a slightly more stable distribution. To further distinguish between these candidates, we examine the retain unseen ISM in Figure~\ref{fig:uce_alphae_retain_ism}. Here, \(\alpha_e=20\) shows a marginally higher median, indicating slightly better preservation of non-target identities. Based on this trade-off, we select \(\alpha_e=20\) as the most balanced choice.

We also observe that \(\alpha_e\) has negligible impact on KD and eDIFFIQA, both on the forget and retain sets, and we therefore omit those plots for clarity.

\begin{figure}[ht]
    \centering
    \includegraphics[width=0.8\linewidth]{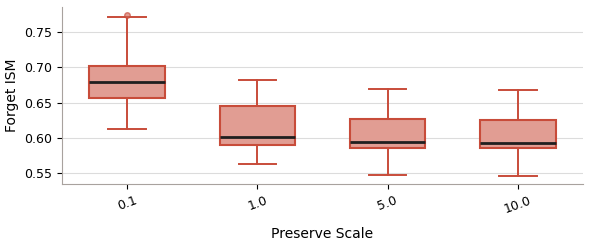}
    \caption{Forget ISM obtained across the grid search for different values of \(\alpha_p\). Each point corresponds to one experiment.}
    \label{fig:uce_alphap_forget_ism}
\end{figure}

We next study the preservation coefficient \(\alpha_p\), for which the behavior is more clearly differentiated. In Figure~\ref{fig:uce_alphap_forget_ism}, the median forget ISM is clearly higher for \(\alpha_p=0.1\), and the distribution is also substantially wider, indicating greater variability across configurations. By contrast, for \(\alpha_p \geq 1\), the forget ISM becomes both lower and more stable. A plausible interpretation is that, when \(\alpha_p\) is too small, the retain identities provide too little structure for the closed-form update, so the edit is insufficiently constrained relative to the surrounding identity geometry. As a result, the displacement of the forget identity is weaker and less consistent, leading to higher forget ISM.

\begin{figure}[ht]
    \centering
    \includegraphics[width=0.8\linewidth]{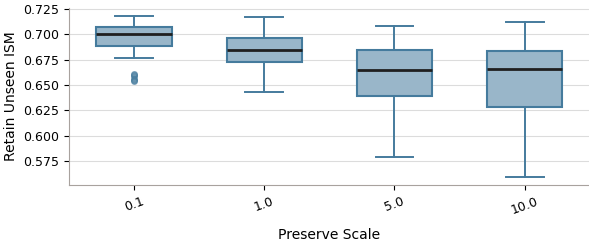}
    \caption{Retain unseen ISM obtained across the grid search for different values of \(\alpha_p\). Each point corresponds to one experiment.}
    \label{fig:uce_alphap_retain_ism}
\end{figure}

Figure~\ref{fig:uce_alphap_retain_ism} further helps distinguish among the remaining candidates. Although the lowest median retain unseen ISM is attained for \(\alpha_p \in \{5,10\}\), the value for \(\alpha_p=1\) remains very high and is accompanied by a noticeably tighter distribution. This suggests that \(\alpha_p=1\) achieves nearly the same level of preservation while being more stable across configurations. Taken together, these trends indicate that \(\alpha_p=1\) provides the best trade-off between effective editing and preservation.

\begin{table*}[ht]
\centering
\small
\resizebox{\textwidth}{!}{
\begin{tabular}{l|c|c|c|c|c|c|c}
\hline
Method
& Forget ISM $\downarrow$
& Retain Unseen ISM $\uparrow$
& SRK $\uparrow$
& $\Delta$ Forget KD $\downarrow$
& $\Delta$ Retain Unseen KD $\downarrow$
& Forget eDIFFIQA $\uparrow$
& Retain Unseen eDIFFIQA $\uparrow$ \\
\hline
Baseline & 0.775 & 0.726 & 0.990 & -- & -- & 0.763 & 0.770 \\
\hline
UCE (\(\alpha_e=20,\alpha_p=1,\lambda=0.7\)) 
& 0.621
& 0.694	
& 0.990
& 4.195
& 1.110	
& 0.749	
& 0.768 \\
\hline
\end{tabular}
}
\caption{UCE reproduction on Arc2Face using the best hyperparameter configuration found in our grid search, \(\alpha_e=20\), \(\alpha_p=1\), and \(\lambda=0.7\). The baseline corresponds to the original Arc2Face model before editing. We report identity-removal metrics (ISM and SRK) together with realism and recognizability metrics (KD and eDIFFIQA).}
\label{tab:uce_main_results}
\end{table*}

\begin{figure}[ht]
    \centering
    \includegraphics[width=0.8\linewidth]{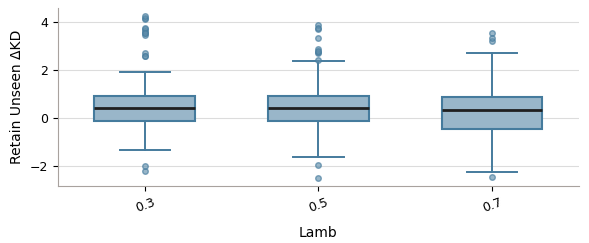}
    \caption{Retain unseen KD obtained across the grid search for different values of \(\lambda\). Lower values indicate better realism.}
    \label{fig:uce_lambda_forget_kd}
\end{figure}

\begin{figure}[ht]
    \centering
    \includegraphics[width=0.8\linewidth]{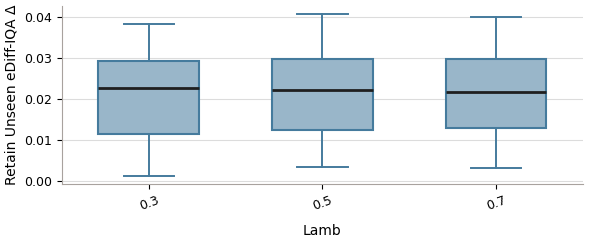}
    \caption{Retain unseen eDIFFIQA obtained across the grid search for different values of \(\lambda\). Higher values indicate better facial quality and recognizability.}
    \label{fig:uce_lambda_forget_ediffiqa}
\end{figure}

Finally, we consider the regularization coefficient \(\lambda\). In this case, the effect on ISM is comparatively limited, so the most informative differences appear in the generation-quality metrics. In particular, Figures~\ref{fig:uce_lambda_forget_kd} and~\ref{fig:uce_lambda_forget_ediffiqa} show a slight but consistent advantage for \(\lambda=0.7\), which tends to yield lower ($\Delta$) KD values and therefore somewhat better realism. At the same time, this value does not noticeably harm the identity-related metrics, nor does it reduce eDIFFIQA in a problematic way. More broadly, one should note that UCE in our setting is not a particularly aggressive method in terms of breaking facial recognizability: eDIFFIQA remains relatively high across the grid search, which is consistent with the qualitative observation that the generated faces remain recognizable even when the identity shift is limited. Based on these trends, we select \(\lambda=0.7\) for the final reported configuration.

Combining the observations above, the best hyperparameter setting is
\[
\alpha_e = 20, \qquad \alpha_p = 1, \qquad \lambda = 0.7.
\]
Table~\ref{tab:uce_main_results} shows the corresponding results, averaged across the forget identities, and compares them with the baseline Arc2Face model. As already suggested by the grid search, UCE slightly lowers forget ISM while keeping good visual quality, but it does not achieve meaningful unlearning under the SRK metric, which stays almost unchanged. This suggests that, in our setting, the closed-form edit of UCE is too weak to produce strong identity removal.

\begin{figure}[t]
        \centering
        \includegraphics[width=0.9\linewidth]{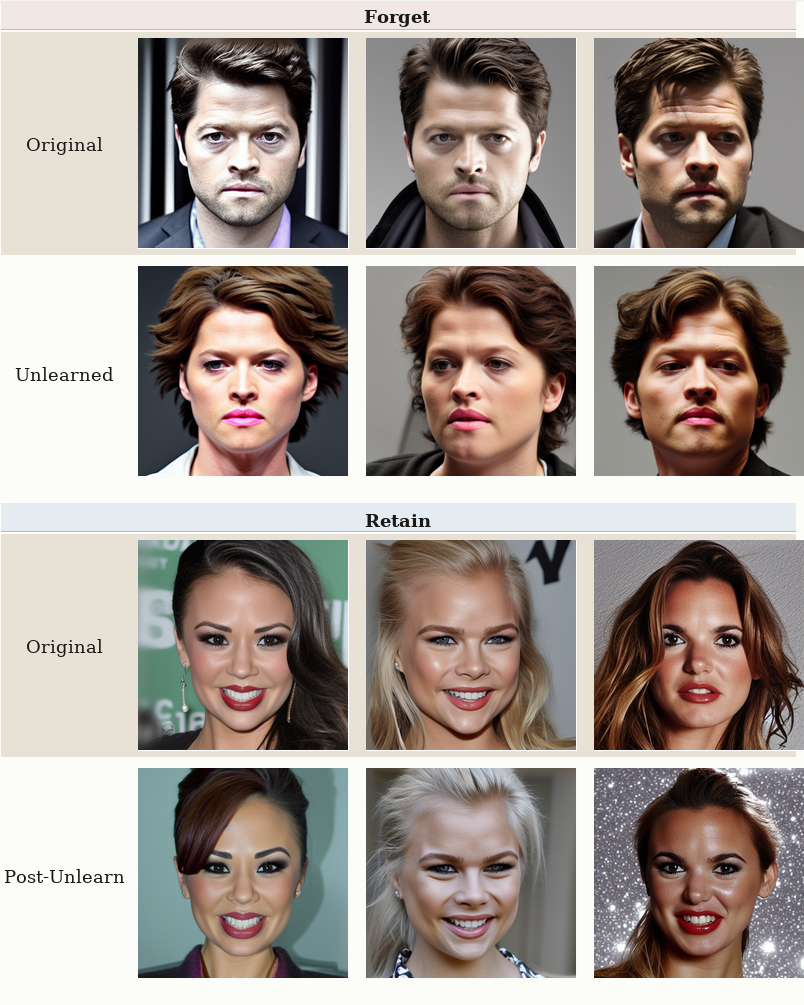}
        \caption{Qualitative results of the UCE reproduction on Arc2Face, using the best configuration found in our grid search: \(\alpha_e = 20\), \(\alpha_p = 1\), and \(\lambda = 0.7\). Rows 1--2 show baseline and edited generations for the forget identity, while Rows 3--4 show the corresponding generations for identities in the unseen retain validation set. The edit induces visible changes, especially on the forget identity, but these are not sufficient to alter its defining attributes in a consistent way, which agrees with the quantitative SRK results.}
        \label{fig:uce_qualitative_results}
\end{figure}

A qualitative example with this configuration is shown in Figure~\ref{fig:uce_qualitative_results}. Some visible changes can be seen on the forget identity, and the generations sometimes move slightly toward the anchor. However, these changes are not strong or consistent enough to modify the main identity traits of the subject. Instead, the edit mostly affects more superficial visual aspects, while the main identity remain. This matches the quantitative results and supports the conclusion that, although UCE is efficient and can produce visible edits, its closed-form strategy is not strong enough for robust identity unlearning in Arc2Face.

\subsection{Unlearn and Protect With ID-guidance}
\label{subsec:wid_reproduction}

This baseline is particularly important in our setting because, to the best of our knowledge, the Unlearn and Protect variant With ID-guidance (WID)~\cite{shaheryar2025unlearn}, together with BIA~\cite{shaheryar2025black}, is one of the very few works that address identity unlearning in diffusion models. Among them, WID is the only prior method specifically formulated for an identity-conditioned diffusion model, which makes it the most direct point of comparison for our approach.
At the same time, WID~\cite{shaheryar2025unlearn} is difficult to reproduce in practice. The original paper introduces a custom identity-conditioned diffusion model based on CosFace identity embeddings, but neither model code nor pretrained weights are publicly available. Although the training losses and some implementation details are described, the exact construction of the diffusion model and conditioning pipeline is not fully specified. In addition, no official implementation of the unlearning procedure is released, and only limited fine-tuning details are reported. In this section, we therefore present our reconstruction of the method, adapted to Arc2Face and integrated into our experimental framework.

\subsubsection{WID Methodology}
WID first builds its own identity-conditioned diffusion model, trained on the CASIA-WebFace dataset, where conditioning is provided by identity embeddings extracted with a pretrained CosFace model. The model is trained with a combination of the standard diffusion denoising objective and an additional identity loss, introduced to preserve facial identity information during reconstruction. This identity-conditioned backbone is then fine-tuned for unlearning.

Following the standard DDPM formulation already introduced in Section~\ref{subsec:preliminaries}, the reverse process defines a conditional generative distribution
\[
p_\theta(x_{0:T} \mid c) = p(x_T) \prod_{t=1}^T p_\theta(x_{t-1} \mid x_t, c),
\]
where here the conditioning variable \(c\) is an identity embedding obtained from CosFace. Given a clean image \(x_0\), the corresponding noisy sample at timestep \(t\) is generated through the usual forward process as \(x_t = \sqrt{\bar{\alpha}_t}\,x_0 + \sqrt{1-\bar{\alpha}_t}\,\epsilon\), with \(\epsilon \sim \mathcal{N}(0,I)\). The main idea of the method is to remove a forget identity \(c_f\) by aligning the conditional distribution of the fine-tuned model under \(c_f\) with that of a frozen pretrained model under a target or anchor identity \(c_a\). Formally, this is expressed through the KL minimization objective
\begin{equation}
\min_{\theta} \;
\mathrm{KL}\Big(
p_{\theta^*}(x_{0:T} \mid c_a)
\;\|\;
p_{\theta}(x_{0:T} \mid c_f)
\Big),
\label{eq:wid_kl}
\end{equation}
where \(\theta^*\) denotes the frozen pretrained parameters and \(\theta\) the fine-tuned ones.

As discussed in the original paper, directly optimizing this KL divergence is computationally challenging, since it would require handling full diffusion trajectories. For this reason, the authors derive an equivalent objective at the denoising level using the standard noise-prediction parameterization \(\epsilon_\theta(x_t,t,c)\). In our notation, their main unlearning loss can be written as
\begin{equation}
\mathcal{L}_{\text{model}}
=
\mathbb{E}_{x_0,\,\epsilon,\,t}
\left[
\left\|
\epsilon_{\theta^*}(x_t, t, c_a)
-
\epsilon_{\theta}(x_t, t, c_f)
\right\|_2^2
\right],
\label{eq:wid_model}
\end{equation}
which enforces that the fine-tuned model, when queried with the forget identity \(c_f\), matches the denoising prediction of the frozen model under the anchor identity \(c_a\).

In addition, the method incorporates a second objective, called \emph{identity loss}, to encourage identity consistency in the reconstructed samples. This loss is computed in the CosFace embedding space and is defined as
\begin{equation}
\mathcal{L}_{\text{id}}
=
1 - \cos\big( \mathrm{CosFace}(x_0), \mathrm{CosFace}(\hat{x}_0) \big),
\label{eq:wid_id}
\end{equation}
where \(\hat{x}_0\) denotes the reconstructed image predicted by the denoising model. This term operates at the image level and encourages the generated sample to remain aligned with the desired identity representation. The final training objective is then given by
\begin{equation}
\mathcal{L}_{\text{total}}
=
\mathcal{L}_{\text{model}}
+
\lambda \mathcal{L}_{\text{id}},
\label{eq:wid_total}
\end{equation}
where \(\lambda\) controls the trade-off between the anchor-guided replacement objective and identity consistency in the reconstructed outputs.
For clarity, we refer to this method as \emph{WID} (with-identity-loss), following the terminology used in Table~1 of the original paper, where the variant incorporating the identity loss is denoted in this way.

\subsubsection{Adapting WID to Arc2Face}

To adapt this method to our setting, we replace the original identity-conditioned backbone with Arc2Face. As a result, identity conditioning is no longer based on CosFace embeddings, but on ArcFace embeddings.

We follow the same anchor-guided principle as in Eq.~\eqref{eq:wid_model}, but reformulate it in the PIU framework. Given a forget identity condition \(c_f\), an anchor identity condition \(c_a\), and a noisy latent \(z_t\), the unlearning objective at the noise-prediction level is
\begin{equation}
\hat{\epsilon}_{\mathrm{forget}}
=
\epsilon_{\theta^*}(z_t,t,c_a)
-
\eta\Big[
\epsilon_{\theta^*}(z_t,t,c_f)
-
\epsilon_{\theta^*}(z_t,t,c_a)
\Big].
\label{eq:wid_adapt_forget_target}
\end{equation}
Equivalently, this can be rewritten as
\begin{equation}
\hat{\epsilon}_{\mathrm{forget}}
=
(1+\eta)\,\epsilon_{\theta^*}(z_t,t,c_a)
-
\eta\,\epsilon_{\theta^*}(z_t,t,c_f),
\label{eq:wid_adapt_forget_target_expanded}
\end{equation}
and therefore the forget objective is
\begin{equation}
\mathcal{L}_{\mathrm{forget}}
=
\mathbb{E}_{z,\epsilon,t}
\left[
\left\|
\epsilon_{\theta}(z_t,t,c_f)
-
\hat{\epsilon}_{\mathrm{forget}}
\right\|_2^2
\right].
\label{eq:wid_adapt_forget_loss}
\end{equation}
If we now fix \(\eta=0\), then
\begin{equation}
\hat{\epsilon}_{\mathrm{forget}}
=
\epsilon_{\theta^*}(z_t,t,c_a),
\label{eq:wid_adapt_eta_zero_target}
\end{equation}
so that the forget loss reduces to
\begin{equation}
\mathcal{L}_{\mathrm{forget}}
=
\mathbb{E}_{z,\epsilon,t}
\left[
\left\|
\epsilon_{\theta}(z_t,t,c_f)
-
\epsilon_{\theta^*}(z_t,t,c_a)
\right\|_2^2
\right].
\label{eq:wid_adapt_eta_zero_loss}
\end{equation}
Hence, the model-alignment loss \(\mathcal{L}_{\mathrm{model}}\) used in WID ( Eq.~\eqref{eq:wid_model}), corresponds exactly to the \(\eta=0\) special case of our forget loss. In this sense, NIL (the No-Identity-Loss version of \cite{shaheryar2025unlearn}) can be interpreted as a restricted instance of our anchor-guided formulation in which the forget trajectory is matched directly to the frozen prediction under the anchor identity, without the additional repulsive guidance induced by \(\eta>0\).

\begin{table*}[ht]
\centering
\small
\begin{tabular}{cc|ccccc}
\hline
LR & \(\lambda\)
& Forget ISM \(\downarrow\)
& Retain ISM \(\uparrow\)
& SRK \(\uparrow\)
& \(\Delta\)KD Forget \(\downarrow\)
& \(\Delta\)KD Retain \(\downarrow\) \\

\hline
\(5\times 10^{-6}\) & 0.1 & 0.322 & 0.443 & 41.0  & 4.932 & {4.252} \\
\(5\times 10^{-6}\) & 0.3 & 0.316 & 0.446 & 42.0 & 5.545 & 4.766 \\
\(5\times 10^{-6}\) & 0.5 & 0.313 & 0.447 & {43.0} & 5.104 & 4.670 \\
\hline
\(10^{-5}\)         & 0.1 & {0.297} & 0.396 & 31.0 & 4.390 & 5.033 \\
\(10^{-5}\)         & 0.3 & 0.294 & 0.400 & 34.0  & 3.984 & 5.323 \\
\(10^{-5}\)         & 0.5 & 0.290 & 0.396 & 32.0 & 3.777 & 4.837 \\
\hline
\end{tabular}
\caption{Results of the WID adaptation after \(100\) optimization steps for tested combinations of learning rate and identity-loss weight \(\lambda\).}
\label{tab:wid_grid_100}
\end{table*}

The second term in WID is an identity-consistency loss that acts in face-recognition embedding space rather than in diffusion space. In our Arc2Face adaptation, we keep the same idea and define
\begin{equation}
\mathcal{L}_{\mathrm{id}}
=
1-\cos\!\big(s_{\mathrm{id}},\hat{s}_{\mathrm{id}}\big),
\label{eq:wid_adapt_idloss}
\end{equation}
where \(s_{\mathrm{id}}=\mathrm{ArcFace}(x_0)\) denotes the identity embedding of the original clean image, and \(\hat{s}_{\mathrm{id}}=\mathrm{ArcFace}(\hat{x}_0)\) the identity embedding of its reconstruction predicted by the current denoising model. Since the original paper does not specify in detail how \(\hat{x}_0\) is obtained, in our implementation we recover it from the current noise prediction using the standard DDPM relation
\begin{equation}
\hat{x}_0
=
\frac{x_t-\sqrt{1-\bar{\alpha}_t}\,\epsilon_\theta(x_t,t,c_f)}
{\sqrt{\bar{\alpha}_t}}.
\label{eq:wid_x0_pred}
\end{equation}
In Arc2Face, the same relation is applied in latent space, using the noisy latent \(z_t\) and the corresponding predicted noise. The resulting estimate is then decoded to image space, and its ArcFace embedding is used to compute \(\hat{s}_{\mathrm{id}}\).

Although Eq.~\eqref{eq:wid_adapt_idloss} is the conceptual form described in the paper, Algorithm~1 in \cite{shaheryar2025unlearn} suggests an equivalent implementation in terms of squared Euclidean distance between identity embeddings, namely
\begin{equation*}
\mathcal{L}_{\mathrm{id}}
=
\bigl\|
s_{\mathrm{id}}-\hat{s}_{\mathrm{id}}
\bigr\|_2^2.
\label{eq:wid_adapt_idloss_l2}
\end{equation*}
A natural explanation is that, even though this is not stated explicitly, the two forms are equivalent (proportionally) whenever the identity embeddings are \(\ell_2\)-normalized. Indeed, if \(\|u\|_2=\|v\|_2=1\), then
\begin{align*}
\|u-v\|_2^2
&=
(u-v)^\top(u-v) \\
&=
\|u\|_2^2+\|v\|_2^2-2u^\top v \\
&=
2-2u^\top v \\
&=
2\bigl(1-\cos(u,v)\bigr).
\end{align*}
Therefore, for normalized embeddings,
\begin{equation*}
1-\cos(u,v)
=
\frac{1}{2}\|u-v\|_2^2.
\end{equation*}
Hence, both objectives have exactly the same minimizers, since they differ only by a positive constant factor. As in our discussion of the SISS reproduction, such a constant scaling does not change the optimization target, and can be absorbed into the coefficient that controls the strength of the term. Defining \(\lambda\) as the effective weight of the identity-consistency loss, we may therefore write the objective in the equivalent form
\begin{equation}
\mathcal{L}_{\mathrm{total}}
=
\mathcal{L}_{\mathrm{forget}}
+
\lambda
\bigl\|
s_{\mathrm{id}}-\hat{s}_{\mathrm{id}}
\bigr\|_2^2,
\qquad \text{with } \eta=0.
\label{eq:wid_adapt_total}
\end{equation}

Overall, this preserves the original spirit of WID while making it compatible with Arc2Face. Since the forget term coincides exactly with the \(\eta=0\) case of our method, we use the same forget and anchor identities introduced in Section~\ref{subsec:surgical_layers}. Moreover, because WID does not include a dedicated study of anchor selection, we keep our main anchor-selection strategy, which remains well motivated by the ArcFace embedding-space geometry discussed in Section~\ref{subsec:threshold_optimized_anchor_selection}.

\subsubsection{WID Experiments}

The general setup follows Section~\ref{sec:experiments}. A key difference from our method is that this adaptation requires real image latents to evaluate the identity-consistency term \(\mathcal{L}_{\mathrm{id}}\). In particular, the loss~\ref{eq:wid_adapt_idloss} is computed from a reconstruction \(\hat{x}_0\), whose ArcFace embedding is compared against that of the original image. To obtain \(\hat{x}_0\), we use Eq.~\eqref{eq:wid_x0_pred} with the standard DDPM scheduler employed in Arc2Face.

This makes the method substantially more expensive than the other baselines reproduced in our study, since each optimization step additionally involves reconstructing \(\hat{x}_0\), decoding it to image space, and extracting ArcFace features. In our implementation, this results in training that is approximately seven times slower than PIU, which considerably limits the feasible hyperparameter budget.

\begin{table*}[ht]
\centering
\small
\resizebox{\textwidth}{!}{
\begin{tabular}{l|c|c|c|c|c|c|c}
\hline
Method
& Forget ISM $\downarrow$
& Retain Unseen ISM $\uparrow$
& SRK $\uparrow$
& $\Delta$ Forget KD $\downarrow$
& $\Delta$ Retain Unseen KD $\downarrow$
& Forget eDIFFIQA $\uparrow$
& Retain Unseen eDIFFIQA $\uparrow$ \\
\hline
Baseline & 0.775 & 0.726 & 0.99 & -- & -- & 0.763 & 0.770 \\
\hline
WID
& 0.322
& 0.443	
& 41.0
& 4.932 & 4.252
& 0.732	
& 0.749 \\
\hline
\end{tabular}
    }
\caption{WID reproduction on Arc2Face using the best hyperparameter configuration found in our grid search, lr = $5e^{-6}$, \(\lambda=0.1\), and steps = $100$. The baseline corresponds to the original Arc2Face model before editing. We report identity-removal metrics (ISM and SRK) together with realism and recognizability metrics (KD and eDIFFIQA).}
\label{tab:wid_final_results}
\end{table*}

For this reason, we restrict the exploration to a small grid over the learning rate and the identity-loss weight \(\lambda\). The original paper reports a learning rate of \(10^{-4}\), but in our Arc2Face adaptation we found this value to be unstable, leading to a clear degradation of the model. We therefore evaluate the smaller learning rates \(5\times 10^{-6}\) and \(10^{-5}\). The former is taken as a conservative setting, while the latter is included to test whether a stronger optimization regime can compensate for the reduced training budget without destabilizing training. For the identity-consistency weight, we consider \(\lambda \in \{0.1, 0.3, 0.5\}\). This includes the default value \(\lambda=0.3\) used in the original work, while also allowing us to assess the sensitivity of the method to weaker or stronger identity preservation.

Finally, while the original paper recommends between \(1000\) and \(2000\) optimization steps, such a budget is infeasible here. We therefore train for \(300\) steps and additionally evaluate checkpoints at steps \(100\) and \(200\), in order to assess more carefully how much optimization is actually needed in Arc2Face. The remaining settings are kept fixed: batch size \(64\), AdamW optimizer, and weight decay \(0\).

\subsubsection{WID Results and Discussion}

We first study the effect of the number of optimization steps, comparing \(100\), \(200\), and \(300\) updates. We observe a clear and consistent pattern: increasing the number of steps degrades both unlearning and generation quality. This trend is already evident from SRK, which is the main unlearning metric used throughout our analysis. For the learning rate \(5\times 10^{-6}\), the best SRK value across the tested \(\lambda\) values is \(43\) at step \(100\), \(38\) at step \(200\), and \(31\) at step \(300\). A similar behavior is found for the learning rate \(10^{-5}\), where the corresponding best SRK values are \(34\), \(28\), and \(31\). The same tendency also appears in the other metrics, so we do not report the full step-by-step comparison here. Overall, these results provide sufficient evidence to select \(100\) steps as the best training horizon for this adaptation.

We therefore focus the main comparison on the \(100\)-step checkpoints. Table~\ref{tab:wid_grid_100} reports the main metrics of interest for all tested combinations of learning rate and identity-loss weight \(\lambda\).

From Table~\ref{tab:wid_grid_100}, a clear pattern emerges regarding the learning rate. Although \(10^{-5}\) produces a somewhat stronger effect on forget ISM, this does not translate into better SRK, because retain identities deteriorate much more quickly. In contrast, the smaller learning rate \(5\times 10^{-6}\) is less aggressive and preserves the rest of the identity space more effectively, which leads to better overall SRK. The same trade-off is also visible in the quality metrics: while \(10^{-5}\) can slightly improve \(\Delta\)KD on the forget set, this gain is offset by a worse \(\Delta\)KD on the retain set. For this reason, we prefer \(5\times 10^{-6}\), which achieves a better balance between forget strength and preservation of non-target identities.

The influence of \(\lambda\) is comparatively weaker on the unlearning metrics. Table~\ref{tab:wid_grid_100} reports also the forget and retain ISM values across the tested \(\lambda\) values for the preferred learning rate \(5\times 10^{-6}\). While the differences in ISM are relatively small, the quality metrics show a clearer dependence on \(\lambda\). In particular, \(\lambda=0.1\) yields substantially smaller distribution shifts in both forget and retain \(\Delta\)KD, indicating better realism and overall post-unlearning quality. We therefore select \(\lambda=0.1\) as the best operating point.

Based on this analysis, our final selected configuration is \(\lambda=0.1\), learning rate \(5\times 10^{-6}\), and \(100\) optimization steps. The final comparison against the baseline model is reported in Table~\ref{tab:wid_final_results}.

Overall, the results are promising but also reveal an important limitation of the method. On the positive side, the forget ISM decreases noticeably, indicating that the target identity is effectively altered, while image quality remains reasonably high. However, the main weakness of the method lies on the retain side. Since WID does not include an explicit preservation term over non-target identities, it cannot maintain them as effectively as our method. As a result, retain metrics deteriorate and the overall identity preservation of the model is reduced.

\begin{figure}[t]
        \centering
        \includegraphics[width=0.9\linewidth]{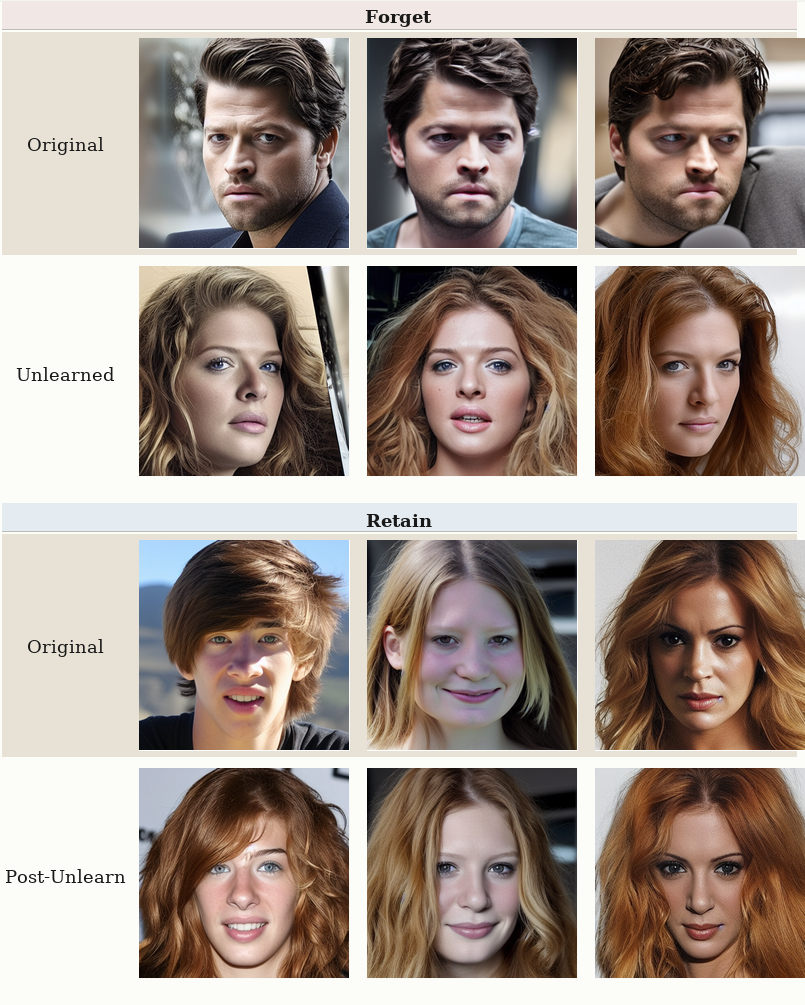}
        \caption{Qualitative results of the WID adaptation on Arc2Face using the selected configuration: learning rate \(5\times 10^{-6}\), \(\lambda=0.1\), and \(100\) optimization steps. Row 1 shows baseline generations for the forget identity, and Row 2 the corresponding generations after unlearning. Row 3 shows baseline generations for identities in the unseen retain validation set, and Row 4 their generations after unlearning. The method effectively alters the forget identity while largely preserving visual quality, but it also changes non-target identities, which is consistent with the drop observed in the retain metrics.}
        \label{fig:wid_qualitative_results}
\end{figure}

This behavior is also visible qualitatively in Figure~\ref{fig:wid_qualitative_results}. On the forget side, generations remain visually plausible and high-quality, and the target identity no longer appears to correspond to the original person. The more problematic effect appears on the retain side: although the images do not collapse in quality as severely as in some other baselines, the generated faces are no longer as faithful to their original identities. This suggests that the update does not only move the forget identity away from its original region, but also shifts part of the surrounding identity distribution toward the anchor subspace.

% \clearpage
% \fi

\end{document}